\documentclass[10pt,journal,compsoc]{IEEEtran}
\usepackage{graphicx}
\usepackage{picture}
\usepackage{float}
\usepackage[table,xcdraw]{xcolor}
\usepackage{multirow}
\usepackage{makecell}

\ifCLASSOPTIONcompsoc
\usepackage[caption=false,font=normalsize,labelfont=sf,textfont=sf]{subfig}
\else
\usepackage[caption=false,font=footnotesize]{subfig}
\fi

\usepackage{caption} 
\ifCLASSOPTIONcompsoc
  \usepackage[nocompress]{cite}
\else
  \usepackage{cite}
\fi

\usepackage[ruled,vlined]{algorithm2e}

\usepackage{listings} 
\usepackage{amsmath}  
\usepackage{version}

\usepackage{lscape} 
\usepackage{nccmath} 
\usepackage{amsfonts}

\usepackage[utf8]{inputenc}

\usepackage{xcolor}

\newcommand{\CHANGE}[1]{\color{black}{#1}}

\definecolor{dartmouthgreen}{rgb}{0.05, 0.5, 0.06}
\newcommand{\NEWCHANGE}[1]{\color{black}{#1}}

\begin{document}

\title{Increasing the Confidence of Deep\protect\\Neural Networks
by Coverage Analysis}

\author{Giulio~Rossolini, ~Alessandro~Biondi,~\IEEEmembership{Member,~IEEE},~ 
        and~Giorgio~Buttazzo,~\IEEEmembership{Fellow,~IEEE}
\IEEEcompsocitemizethanks{\IEEEcompsocthanksitem {G. Rossolini, A. Biondi and G. Buttazzo are with the Department of Excellence in Robotics \& AI, Scuola Superiore Sant'Anna \protect\\
mail:giulio.rossolini@santannapisa.it; ~alessandro.biondi@santannapisa.it; ~giorgio.buttazzo@santannapisa.it}
\IEEEcompsocthanksitem{This work has been submitted to the IEEE for possible publication. Copyright may be transferred without notice, after which this version may no longer be accessible.}
}}

\IEEEtitleabstractindextext{%
\begin{abstract}

The great performance of machine learning algorithms and deep neural networks in several perception and control tasks is pushing the industry to adopt such technologies in safety-critical applications, as autonomous robots and self-driving vehicles. At present, however, several issues need to be solved to make deep learning methods more trustworthy, predictable, safe, and secure against adversarial attacks. Although several methods have been proposed to improve the trustworthiness of deep neural networks, most of them are tailored for specific classes of adversarial examples, hence failing to detect other corner cases or unsafe inputs that heavily deviate from the training samples.

This paper presents a lightweight monitoring architecture based on coverage paradigms to enhance the model robustness against different unsafe inputs. 
In particular, four coverage analysis methods are proposed and tested in the architecture for evaluating multiple detection logic.
Experimental results show that the proposed approach is effective in detecting both powerful adversarial examples and out-of-distribution inputs, introducing limited extra-execution time and memory requirements.

\end{abstract}

\begin{IEEEkeywords}
Neural Networks Coverage, DNNs Robustness, Adversarial Examples Detection
\end{IEEEkeywords}
}

\maketitle

\IEEEdisplaynontitleabstractindextext

%
\IEEEpeerreviewmaketitle

\section{Introduction}

Recent developments of machine learning algorithms exhibited superhuman performance to solve specific problems, as image classification, object detection, control, and strategy games.
However, most of the AI algorithms developed today have been used for non-critical applications, as face aging, speech recognition, text prediction, gaming, image restoration and colorization, etc. 
Due to their excellent performance, there is a great industrial interest in using deep neural networks (DNNs) and, more in general, machine learning algorithms in autonomous systems, as robots and self-driving vehicles.
When moving to such safety-critical application domains, several questions arise: can we trust machine-learning algorithms as they are? Are they prone to cyber-attacks? What to do if they fail? Are outputs generated within bounded response times? To address these questions, several issues need to be addressed at different levels of the architecture, as security, safety, explainability, and predictability. 

This paper focuses on security and safety, which are quite intertwined.
Several works have shown that DNN models are quite sensitive to small input variations, which can cause a DNN to produce a wrong prediction. This property has been used to generate adversarial examples (AEs), which are specially crafted inputs that appear genuine to humans but are incorrectly classified by the model with a high confidence.

Another serious problem is caused by inputs that are significantly different from those used to train the model. In such corner cases, the output of the model cannot be trusted. The problem can arise for several reasons.
A common situation occurs when the training set contains a bias, introduced when most objects of a class appear on a similar background. In this case, the model can learn such irrelevant features, generating a wrong prediction with high confidence when an object belonging to a different class appears on the same background.

Another problem arises from the fact that, normally, DNNs for classification tasks are trained to classify $m$ given classes using data sets containing only objects belonging to those classes. In this case, in the presence of a new object that does not belong to any of the $m$ classes, the network cannot say “I don’t know”, and could be forced to take a decision by the softmax output layer, which imposes that the sum of the output values has to be 1.0.
In general, when the feature distribution of the input is significantly different from those that characterize the data used to train the model, the output should not be trusted, because the input could activate unusual neural paths that could cause a wrong prediction with high confidence.

To improve the robustness of DDNs against adversarial attacks, several methods have been proposed in the literature. Some of these methods consist of augmenting the training data by suitable transformations, or adversarial examples. Other works enforced the robustness of DNNs by defenses applied at inference or testing time. They are reviewed in more detail in Section \ref{s:relatedwork}.

{\CHANGE 

\textbf{Contribution.} Inspired by coverage metrics for DNNs studied in previous work for offline testing purposes, 
this work proposes new methods to enhance the trustworthiness of DNNs by providing a confidence value coupled to the prediction made by the network. The confidence value is obtained by coverage-based run-time monitoring.

In particular, while the prediction score of DNNs is still associated with the model softmax output, the prediction confidence is computed by analyzing the activation state of a large number of internal neurons and comparing it with the corresponding state produced from a trusted data set.
The proposed approach consists in two phases.
In a preliminary (off-line) phase, the neuron outputs, in each layer and for each class, are analyzed and aggregated into a set of covered states, which all together represent a \textit{signature} that describes how the network responded to those samples for each given class.
Such a signature is derived using a trusted dataset, that is, a subset of the training samples for which that DNN generates a correct prediction with a softmax probability higher than a given threshold.
Then, at runtime, each new input is subject to an evaluation phase, in which the activation state produced by the input in each layer is compared with the corresponding signature for the class predicted by the network.
The approach is schematically illustrated in Figure~\ref{fig:idea}.

\begin{figure}[tp]
\centering
\makebox[\columnwidth]{\includegraphics[scale=0.7]{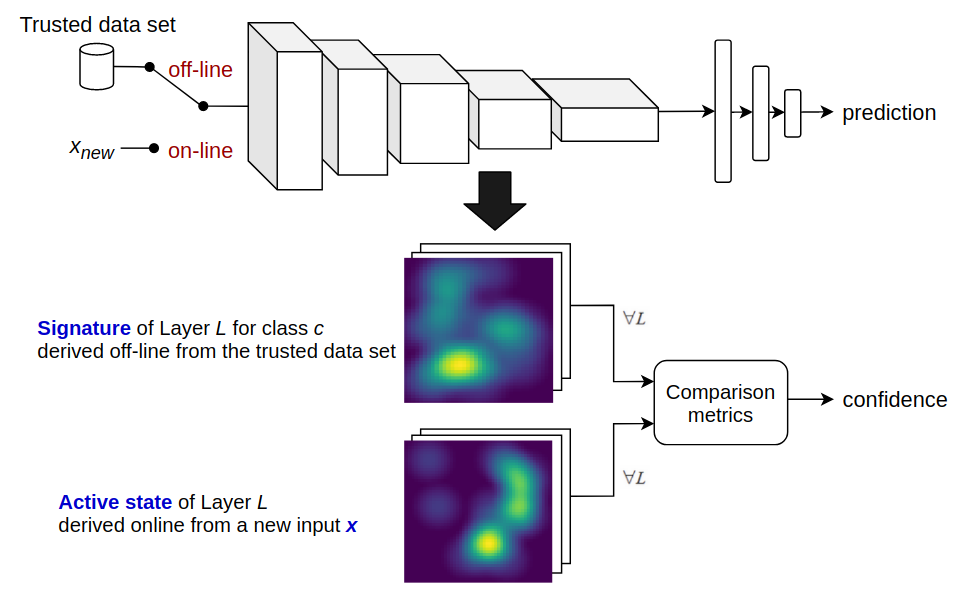}}
\caption{\small Simplified scheme of the approach used to derive a confidence level for each prediction.}
\label{fig:idea}
\vspace{-1.5em}
\end{figure}

The adopted coverage methods are procedural, well-defined, and not based on learning-enabled components: as such, they provide solid foundations for checking the run-time behavior of DNNs in critical systems. 
The proposed approach can be used for detecting adversarial examples and, more in general, inputs that do not comply with the distribution of a pre-selected set of inputs used as a reference trusted data set. 
While many previous works rely on input transformations to accomplish the later tasks, this paper performs run-time monitoring by exploiting a white-box model analysis.
Furthermore, the proposed approach proved to be suitable for resource-constrained devices, since it introduces limited extra latency at inference time and a modest memory footprint in most cases. 

In summary, this paper provides the following contributions:
\begin{itemize}
\item It introduces a novel and flexible monitoring architecture for enhancing the trustworthiness of DNNs through coverage criteria.

\item It presents four Coverage Analysis Methods (CAMs) to instantiate the proposed monitoring architecture. 

\item It presents an implementation of the proposed monitoring architecture and the four CAMs as GPU-accelerated extensions of the Caffe framework.

\item It evaluates the proposed methods in terms of detection performance, additional inference time, and memory footprint.
State-of-the-art methods to generate unsafe inputs have been used for the evaluation. 
\end{itemize}

\textbf{Paper structure.}
The rest of the paper is organized as follows. Section~\ref{s:relatedwork} reviews the state-of-the-art methods focused on improving the robustness of DNNs. 
Section~\ref{s:idea} introduces the notation adopted in the paper.
Section~\ref{s:monitor} describes the proposed architecture with more details.
Section~\ref{s:CAMs} formally presents four CAMs and the corresponding algorithms.
Section~\ref{s:impl} illustrates the implementation of the proposed approach in Caffe.
Section~\ref{s:exps} reports the experimental results. Finally, Section~\ref{s:conc} states the conclusions and future work.
}

\section{Related work} \label{s:relatedwork}
The literature related to this work is quite vast and can be classified into (i) methods based on data augmentation,
(ii) inference- and testing-time defense mechanisms, and (iii) methods to quantify the trustworthiness of DNNs.

\textbf{Data augmentation.}
To improve the robustness of DNNs against adversarial attacks (e.g.,~\cite{goodfellow2014explaining},~\cite{BIM},~\cite{papernot_norm1},
~\cite{CarliniAttack}), various methodologies have been proposed in the literature.
Most of them rely on data augmentation and work by enriching the training set with new samples 
generated via transformations of the available training inputs.
This approach has been followed by Kurakin et al.~\cite{KurakinGB16a}, Pei et al.~\cite{DeepTest}, and Shaham et al.~\cite{Shaham_2018}.
Data augmentation has shown to significantly improve robustness of neural networks and their capability to generalize among new valid inputs. However, it does not help defend against unseen AEs. 

Alternative approaches, such as those proposed by Sinha et al.~\cite{SinhaND18} and Tramèr et al.~\cite{conf/iclr/TramerKPGBM18}, 
tried to make DNNs more robust by performing a specific type of data augmentation called adversarial training.
Adversarial training is a type of defense, first proposed by Goodfellow et al.~\cite{goodfellow2014explaining}, that works by enriching the training set with pre-computed AEs and retraining the network until it learns to classify them correctly.  
Papernot et al.~\cite{Distillation_Papernot} have instead extended a DNN distillation technique \cite{hinton2015distilling} to devise a new training method capable of improving the network robustness with respect to AEs.
Despite these interesting works, it is becoming more and more evident that avoiding unsafe behaviors of DNNs is a very complex or even impossible task~\cite{adv_inevitable},~\cite{carlini_detection_problem}.
Therefore, it is preferable to accept the existence of unsafe inputs and focus on their detection by means of defense mechanisms.

\textbf{Inference- and testing-time defenses.}
Recent works tried to address these issues with new approaches based on enforcing the robustness of DNNs by means of defenses applied at inference or testing time.
They rely on an online validation of the DNN behavior to establish whether a new input is likely to be dangerous. 
Most of the proposed detection methods are based on the observation that AEs typically lie in the proximity of the classification boundaries~\cite{abasi2018},~\cite{hidden_space}.
Relevant examples of such works are those by Wang et al.~\cite{Wang_2019},
which attempts at validating network inputs using voting techniques on multiple mutations of the network model,
and Srisakaokul et al.~\cite{MULDEF}, which assembles multiple robust network models for improving the output accuracy.
Biggio et al.~\cite{biggio} pointed out the weaknesses of several types of Deep Learning algorithms using gradient-based attacks to question a general detection mechanism, examining different levels of knowledge of the attacked system. 
Also Carlini and Wagner \cite{carlini_detection_problem} addressed these issues by showing the poor effectiveness of various detection approaches. 

{
\NEWCHANGE
Other detection methods, such as Features Squeezing~\cite{feature_squeezing} and Vision-Guard~\cite{vision_guard} study the divergence of the probabilities score vector among network outputs when providing as network inputs multiple versions of the same image obtained with transformations (e.g., jpeg compression, rotation, etc.). 
These methods were conceived by focusing on the fragility of specific types of AEs, 
as also pointed out in~\cite{nesti}.
Other approaches, such as DefenseGan~\cite{defensegan} and MagNet~\cite{conf/ccs/MengC17}, use generative adversarial networks and autoencoders, respectively, to remove possible malicious perturbations from the input sample, which could lead the network to fail the prediction.

A common drawback of several strategies mentioned above is that they are not explicitly conceived to improve the trustworthiness of the network in general, \NEWCHANGE{i.e., by deriving interpretable assessments of the model behavior at run-time}, but rather to counteract specific classes of AEs. 
As such, they may fail in detecting other corner cases or other dangerous inputs (e.g., inputs that are unsafe due to over-generalization \cite{abasi2018} or AEs that are not compliant with a norm-based definition~\cite{survey}~\cite{adv_patch}).
Section~\ref{s:exps} shows how the approaches proposed in this work are capable of improving the detection performance against different types of unsafe inputs, also introducing less run-time overhead.

\textbf{Defenses based on internal network analysis.}
Much closer to this paper are other works that investigated the internal behavior of DNNs to improve the trustworthiness of their outputs.
For instance, Papernot et al.~\cite{k_nearest} introduced a new inference logic for DNN-based classifiers, called DkNN. According to this method, a classical inference is used for extracting the internal activations values of the network. 
Then, such activations are processed with a statistical analysis based on the k-Nearest Neighbors algorithm to decide the
network output.
This approach has been shown to improve the classification performance of DNNs in the presence of malicious inputs, possibly due to the fact that it relies on a more interpretable prediction score than a classical softmax-based output, which has frequently been cited as a poorly credible confidence metrics~ \cite{sensoy_evidential},~\cite{ozbulak_softmax},~\cite{k_nearest}. 

Note that most of the works cited above were not designed to evaluate the trustworthiness of the original DNN model. Rather, they drastically change the inference logic or the input data to improve the robustness of the model. Nevertheless, it has been shown that they can still fail to detect several unsafe inputs~\cite{magnet_attack}, ~\cite{dknn_attack}. 
Furthermore, it is important to remark that most of the works discussed are not suitable for being applied in real-time and/or on embedded devices.
Indeed, they require accessory software components that considerably increase the memory footprint (e.g.,~\cite{k_nearest}) and the running time of the inference phase (e.g.,~\cite{defensegan}).
%

Finally, it is worth mentioning other relevant approaches that improve the trustworthiness of DNNs by processing internal network activations through machine learning algorithms.
Sotgiu et al.~\cite{sotgiu} proposed a complementary architecture to the original DNN, called Deep Neural Rejection (DNR), which quantifies the coherency of the neuron outputs generated by a new input with the respect to the one generated by training samples. Although the basic concept is very close to the DkNN~\cite{k_nearest}, DNR uses Support Vector Machines with Radial Basis Kernels.
Also, Carrara et al.~\cite{carrara} followed a similar approach, using a Long Short-Term Memory (LSTM) to detect unexpected activation patterns deemed as consequences of adversarial examples.
A drawback of these approaches is that the machine learning used to improve the trustworthiness of DNNs may in turn be a source of untrustworthiness for the whole system, hence just shifting the problem.

}

\textbf{Quantification of the DNN trustworthiness.}
Significant attention has also been devoted to metrics that quantify the trustworthiness of DNNs, possibly with the end of matching certification requirements with formal guarantees.
As Huang et al.~\cite{survey} pointed out, it is useful to distinguish between verification and testing methods. The goal of verification methods is to derive formal robustness properties for neural networks ~\cite{convex_verirication}, ~\cite{weng2018evaluating},~ \cite{formalguarantee}. However, due to inherently hardness of the corresponding verification problems, they suffer of scalability issues, especially when applied to modern DNN architectures~\cite{safetyVerification}.
For this reason, testing methods are usually preferred to approximate the certification process to work on a finite set of test inputs. 

The completeness of the number and type of the generated test cases is usually measured by applying \emph{coverage criteria} for DNNs, which 
are considerably different from those that are commonly used for classical software~\cite{Zhu1997SoftwareUT}. 
Pei et al.~\cite{DeepTest} first proposed neuron coverage as a criterion for testing DNNs.
Subsequently, multiple testing metrics based on neurons were proposed~\cite{DeepGauge}. 

In addition to the neuron-based criteria, more accurate criteria based on Modified Condition/Decision Coverage (MC/DC) \cite{mcdc} were proposed by Sun et al.~\cite{mdcddnn1} to improve the coverage performance.
Coverage criteria for DNNs provide a reasonable compromise between formal robustness guarantees and their computational cost.
They are also applied in both white-box, black-box, and concolic testing approaches to extract a large set of AEs for adversarial training~\cite{DeepGauge},~\cite{DeepTest},~\cite{mdcddnn1}.
{\CHANGE 
It is worth noting that such coverage techniques for DNNs have all been conceived for testing purposes and, per se, do not constitute defense mechanisms. 
{\NEWCHANGE Indeed, they are not capable to detect unsafe inputs, even if applied online.}
Conversely, the coverage-based methods proposed in this work are not aimed at retrieving static trustworthiness measures off-line, but at accomplishing a run-time monitoring of the model behaviors against new inputs.
As such, the methods proposed in this work are not comparable with the ones in~\cite{DeepGauge},~\cite{DeepTest},~\cite{mdcddnn1} because they serve a different purpose.

Finally, it is also worth mentioning that several studies \cite{canada_nc, yan_nc} argue how coverage metrics for DNN testing are not correlated to adversarial examples.  In particular, they state a limited relationship between inputs generated to meet coverage criteria and the ones obtained by popular adversarial attacks. Although our work still involves coverage metrics for DNNs, its final objective and the approach are completely different from the ones studied in \cite{canada_nc, yan_nc}, making the corresponding observations not applicable. As a matter of fact, the experimental results presented in Section~\ref{s:exps} demonstrate the capability of our methods of detecting adversarial examples, meaning that it exists a substantial correlation between them and the proposed methods. Furthermore, as supplementary material, we report additional tests to demonstrate the relationship between our approaches and the adversarial examples generated with popular attacks.}

\textbf{This work.}
Differently from previous work, this paper presents a general approach to monitor the network state at inference time (white-box analysis), which is then used to employ new algorithms based on coverage methods to quantify the trustworthiness of the network output as well as detect dangerous inputs on-the-fly when they are presented to the network. Traditional, well-defined procedural algorithms are used for this purpose. The standard inference process of DNNs is retained and extended just to extract intermediate results such as the neuron outputs.
The approach relies on a new kind of layer that runs a passive coverage verification to detect unsafe activation patterns with limited extra execution time.

\section{Terminology and notation} \label{s:idea}
{\CHANGE
Before illustrating the proposed coverage monitoring pipeline, this section introduces the required formalism.
}
A feed-forward DNN is an architecture composed of a set of $K$ \emph{layers} $L = \{L_{k} ~|~ k \in \{1, \ldots, K\} \}$ linked by a series of weighted connections.
Each layer $L_{k}$ consists of a set $N_k$ of $l_{k}$ \emph{neurons}, where the $j$-th neuron is denoted by $n_{k,j} \in N_k$. 
The whole set of neurons (except the input ones) is defined as $N = \{ n_{k,j} ~|~ k \in \{2,\ldots,K\}, j \in \{1, \ldots,l_{k}\} \}$. The neurons within each layer $L_{k}$ are organized into $C_k \geq 1$ \emph{channels}, 
where the set of neurons in the $i$-th channel is denoted by $N^{\text{ch-}i}_k \subseteq N_k$.

Each neuron $n_{k,j} \in N$ is associated with two variables $u_{k,j}$ and $v_{k,j}$ that denote its value before and after the activation function $\phi_{k}$, respectively.  
The neuron output is hence defined as
\begin{equation}
v_{k,j} = \phi_{k}(u_{k,j}) ~~~~\text{with}~~~~ u_{k,j} = b_{k,j} + \sum_{h=1}^{l_{k-1}} w_{k,h,j} \cdot v_{k-1,h},
\end{equation}
where $w_{k,h,j}$ is the \emph{weight} of the connection between neurons $n_{k-1,h}$ and $n_{k,j}$, and $b_{k,j}$ is the \emph{bias} term of neuron $n_{k,j}$. In the following, it is sometimes required to distinguish between the output values of the same neuron when the network is provided with different inputs. To this purpose, the output of each neuron $n_{k,j}$ when the network input is $x$ is denoted by $v_{k,j}(x)$.

This work focuses on classification tasks, where the DNN associates a generic input $x$ to a class $\hat{y}$ belonging to a set of $m = l_K$ classes.
Specifically, the network makes a prediction  $\hat{y}$, which is the index of the neuron in the output layer with the largest value, that is $\hat{y} = \text{argmax}_{1 \leq j \leq m} \{ v_{K,j} \}$,
where $\{1, \ldots, m\}$ is the set of classes and each value $v_{K,j}$ represents the prediction score generated by a softmax output.

To help the presentation of the following results, it is also convenient to introduce a vector notation for the DNN parameters.
Let $V_{k}(x)$ be the vector denoting the activations of the neurons in layer $l_k$ for an input $x$, and let $v_{k,j}(x)$ be the j-th element of $V_k(x)$.
In this way, a DNN can be alternatively expressed as a function $f : \mathbb{R}^{l_{1}} \rightarrow \mathbb{R}^{l_{K}}$ such that $ f(x) = \mathbf{\phi_{K}}(\mathbf{\phi_{K-1}}(...\mathbf{\phi_{2}}(x)))$, where $\mathbf{\phi_{k}}$ is the vector-wise version of the activation function $\phi_{k}$. 
Here, $\mathbb{R}^{l_{1}}$ is the input space (e.g., in the case of image classification, it represents all the possible configurations of pixels of the input image), while $\mathbb{R}^{l_{1}} \times ... \times \mathbb{R}^{l_{K}}$ is the network space, which includes all the output values produced by the neurons in $N$.

This work also considers convolutional neural networks (CNNs)~\cite{CONV}, which can be studied as a special case of DNNs as modeled above.

\section{Monitoring architecture} \label{s:monitor}
This section presents the monitoring architecture proposed to detect unsafe inputs at inference time.
It is based on a novel way of applying coverage criteria. Despite coverage criteria were originally conceived for off-line testing purposes, in our work they are leveraged to identify a series of activation patterns that are deemed safe, because generated by the network on a set of trusted inputs (i.e., those producing a correct prediction with a score higher than a given threshold). 
Then, at runtime, such patterns are compared with those generated by a new input, measuring how they match according to a given confidence metric.

\begin{figure}[tp]
\centering
\makebox[\columnwidth]{\includegraphics[scale=0.4]{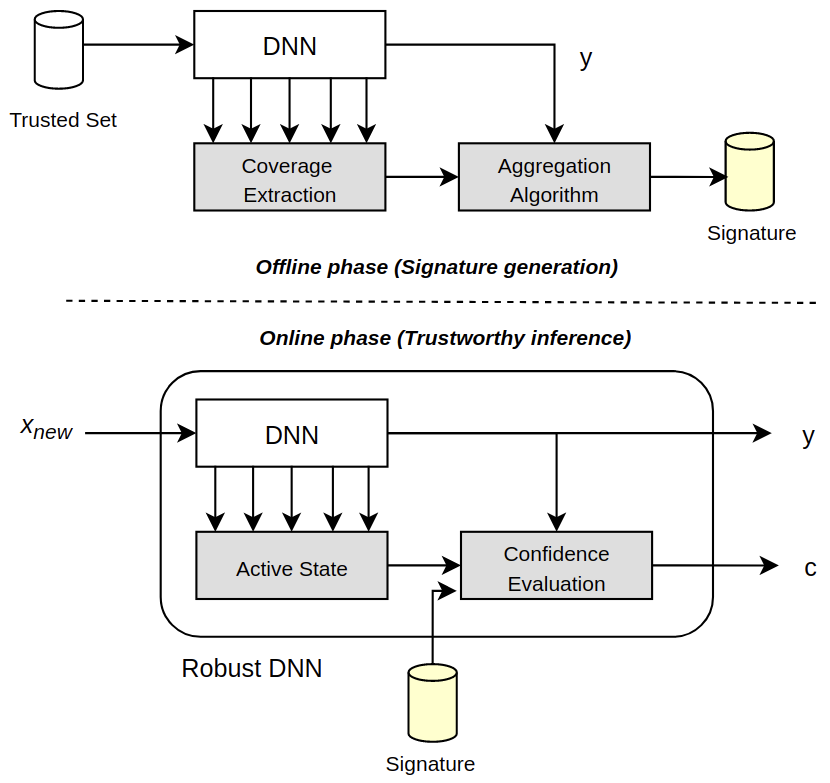}}
\caption{Overview of the monitoring architecture with its offline and online phases.
Grey boxes denote meta-algorithms, i.e., those whose behavior depend on the selected coverage criterion.
}
\label{fig:arch}
\end{figure}

The workflow of the proposed architecture is illustrated in Figure~\ref{fig:arch} and consists of an offline phase (also called \textit{Signature generation}) and an online phase (also called \textit{Trustworthy inference}).

\textbf{Offline phase (\textit{Signature generation}).} 
In this phase, the network processes a set of trusted inputs, denoted by \emph{Trusted Set}, that generate correct network outputs with a high prediction score.
Each sample in the Trusted Set is used to perform inference on the target trained DNN. During inference, the intermediate results produced by the various layers of the DNN (e.g., the neuron outputs) are recorded and used to apply a certain 
coverage criterion, denoted by \emph{Coverage Analysis Method} (CAM) in the following. The coverage results across all inputs in the Trusted Set are then grouped by the corresponding output classes and aggregated to produce a representation of the covered activation patterns (by the Aggregation Algorithm). This representation is then compressed to produce a file called \emph{DNN Signature}.
The DNN Signature encodes all network activation patterns that are deemed safe for each output label, under a certain CAM.
Both the Coverage Extraction and the Aggregation Algorithm depend on the selected CAM.
This work considers four CAMs, which are presented in Section~\ref{s:CAMs}.

\textbf{Online phase (\textit{Trustworthy inference}).}
The network is deployed together with a DNN Signature.
At inference time, given a new input $x_{\text{new}}$, the same CAM used to generate the DNN Signature is applied to extract the coverage result generated by $x_{\text{new}}$, which is referred to as \emph{DNN Active State}.
Given the class $\hat{y}$ predicted for $x_{\text{new}}$, a Confidence Evaluation Algorithm computes the matching level between the Active State stimulated by $x_{new}$ and the trusted one encoded in the DNN Signature corresponding to class $\hat{y}$, producing a confidence value for the prediction. When such a confidence is below a given threshold, the input $x_{new}$ is deemed unsafe. The configuration of such a threshold is addressed in Section 7.
The Confidence Evaluation Algorithm also depends on the selected CAM and four options are presented in Section~\ref{s:CAMs}.

\section{Coverage Analysis Methods} \label{s:CAMs}
{\CHANGE
This section presents three CAMs together with the corresponding algorithms required to instantiate the monitoring architecture illustrated in Figure~\ref{fig:arch}. A fourth CAM, which did not achieve acceptable performance during our experimental evaluation, is presented in the supplementary material. 

Note that all state-of-the-art coverage metrics for DNNs introduced in Section \ref{s:relatedwork} are conceived with a different purpose (i.e., DNN coverage testing). 
Nevertheless, some of them inspired the following CAMs, which are instead designed for being applied to the monitoring scheme proposed in Section \ref{s:monitor}. 
}

\subsection{Single-Range Coverage (SRC)} \label{s:SRC}
This CAM is based on the Neuron Boundary Coverage criterion~\cite{DeepGauge} and works by analyzing the range of output values produced by each neuron. During the offline phase, the minimum and maximum output values produced by the neurons when testing the Trusted Set are recorded to set the range of the ``typical'' behavior of the network when stimulated by trusted inputs.

The Trusted Set $S$ is split into $m$ subsets $S_{1},...,S_{m}$, one for each class, where $S_i$ denotes the set of 
inputs in $S$ belonging to the $i$-th class.
The DNN Signature $\sigma_i$ for the $i$-th class (with $i=1,\ldots, m$) is 
a collection of pairs $\sigma_{i,k,j} = (v_{i,k,j}^\text{min}, v_{i,k,j}^\text{max})$, where $v_{i,k,j}^\text{min}$ and $v_{i,k,j}^\text{max}$ denote
the minimum and maximum output values produced by neuron $n_{k,j}$, respectively, over all inputs in $S_i$.
The Aggregation Algorithm of SRC is summarized in Algorithm~\ref{alg:agg-SRC}.

\newcommand{\algrule}[1][.2pt]{\par\vskip.5\baselineskip\hrule height #1\par\vskip.5\baselineskip}
\begin{algorithm}[htb!]
\SetAlgoLined
 \textbf{Input} Trusted Set $S$, trained DNN \\
 \textbf{Output} DNN Signature $\sigma$ \\
 \algrule
 \For{$S_{i} \in S$} { 
 $\sigma_i = \{ \}$\\
 \For{$n_{k,j} \in N$} {
 $v^\text{min}_{i,k,j} = \min_{x \in S_{i}} \{ v_{k,j}(x) \}$ \\
 $v^\text{max}_{i,k,j} = \max_{x \in S_{i}} \{ v_{k,j}(x) \}$ \\
 $\sigma_{i, k, j} = ( v^\text{min}_{i,k,j}, v^\text{max}_{i,k,j} )$  \\
 Add $\sigma_{i, k, j}$ to $\sigma_i$ \\
 }	
 }
 \Return $\sigma = \{\sigma_1, \ldots, \sigma_{m} \}$ \\
\caption{Aggregation Algorithm of SRC.}
\label{alg:agg-SRC}
\end{algorithm}

During the online phase, given a new input $x_{\text{new}}$ and the corresponding class $\hat{y}$ predicted by the DNN, 
the DNN Signature $\sigma_{\hat{y}}$ is compared against the output values produced by the neurons, denoted as the DNN Active State.
The Confidence Evaluation Algorithm of SRC returns a confidence value that is computed as a function of the number of neurons whose output value $v_{k,j}(x_{\text{new}})$ is outside the range specified by pair $\sigma_{\hat{y}, k, j}$.
The confidence is computed as $c = \exp({- \frac{\eta \cdot ln(2)}{\tau_{\hat{y}}}})$, where $\eta$ is the number of neurons out of range and $\tau_{\hat{y}}$ is a class-dependent parameter, called \textit{threshold}, which tunes the slope of the exponential function (the higher the threshold, the faster the function goes to zero). Note that, if $\eta = \tau_{\hat{y}}$, then $c = 0.5$. The specific values for parameters $\tau_{\hat{y}}$ are set with a calibration procedure presented in Section~\ref{s:exp_calib}.
{\CHANGE
The rationale of this formula is to map the computed coverage cost $\eta$ to a confidence value between $0$ and $1$, to be compliant with typical softmax scores. A smooth exponential function is adopted also because it allows assigning a high confidence when just a few values fall outside the ranges in the signature.
}

This procedure is reported in Algorithm~\ref{alg:safety-eval-SRC}, assuming to monitor all the network's neurons $N$.
In practice, to balance performance with computation time, all these algorithms can also be executed on a subset of $N$ (see Section~\ref{s:impl}).

\begin{algorithm}[htb!]
\SetAlgoLined
 \textbf{Input} input $x_{new}$, DNN Signature $\sigma$, trained DNN, thresholds $\tau_i$ \\
 \textbf{Output} confidence $c$ \\
 \algrule
 $\hat{y} = \text{argmax}_{1\leq j \leq m} \{ f(x_{\text{new}})\}$\\
 $\eta$ = $0$ \\
 \For{$n_{k,j} \in N$} {
 Extract $\sigma_{\hat{y}, k, j}$ from $\sigma_{\hat{y}}$\\
 $ ( v^\text{min}_{\hat{y},k,j}, v^\text{max}_{\hat{y},k,j} ) = \sigma_{\hat{y}, k, j}$ \\
 \If{$v_{k,j}(x_{\text{new}}) \notin [v^\text{min}_{\hat{y},k,j}, v^\text{max}_{\hat{y},k,j}] $}  {
 $\eta$ ++
 }
 }
 \Return $c =  \exp({- \frac{\eta \cdot ln(2)}{\tau_{\hat{y}}}})$ \\
 
 \caption{Confidence Evaluation Algorithm of SRC.}
 \label{alg:safety-eval-SRC}
\end{algorithm}

\subsection{Multi-Range Coverage (MRC)}
This CAM is based on the K-Multi-Section coverage criterion~\cite{DeepGauge} and extends SRC by introducing multiple ranges 
to analyze the output of the neurons. The offline phase works as follows.
First, as for SRC, the minimum and maximum output values produced by the neurons when testing the Trusted Set are recorded.
Then, each output range is evenly split into $Q$ sub-ranges. Finally, for all the inputs of the Trusted Set, the algorithm counts the number of times in which the neuron outputs fall in a given sub-range.

Formally, using the same notation introduced for SRC, for each output class with index $i$, the output range of each neuron $n_{k,j}$ is split into $Q$ sub-ranges of size $\Delta_{i, k, j} = (v_{i,k,j}^\text{max}-v_{i,k,j}^\text{min})/Q$.
For convenience, the
last of such sub-ranges is defined as $[v_{i,k,j}^\text{min}+ (Q-1)\Delta_{i, k, j}, v_{i,k,j}^\text{max}]$, while the others have open right endpoints.

The DNN Signature $\sigma_i$ for the $i$-th class is a collection of tuples $\sigma_{i,k,j} = (v_{i,k,j}^\text{min}, v_{i,k,j}^\text{max}, \lambda_{i,k,j}^1, \lambda_{i,k,j}^2, \ldots, \lambda_{i,k,j}^Q)$, where $\lambda_{i,k,j}^q \in [0, 1]$ (with $q=1, \ldots, Q$) is given by the number of times the output of neuron $n_{k,j}$ falls within the $q$-th sub-range when the network is stimulated by inputs in $S_i$, divided by the cardinality of $S_i$ itself.
The Aggregation Algorithm of MRC, reported in Algorithm~\ref{alg:agg-MRC}, first computes the minimum and maximum output values of the neurons, and then the corresponding occurrencies to produce the DNN Signature.

\begin{algorithm}[htb!]
 \textbf{Input} Trusted Set $S$, trained DNN, number of sections $Q$ \\
 \textbf{Output} DNN Signature $\sigma$ \\
 \algrule
\SetAlgoLined
 \For{$S_{i} \in S$} { 
 $\sigma_i = \{ \}$\\
 \For{$n_{k,j} \in N$} {
 $v^\text{min}_{i,k,j} = \min_{x \in S_{i}} \{ v_{k,j}(x) \}$ \\
 $v^\text{max}_{i,k,j} = \max_{x \in S_{i}} \{ v_{k,j}(x) \}$ \\
 $\Delta_{i, k, j} = (v_{i,k,j}^\text{max}-v_{i,k,j}^\text{min})/Q$\\
 $\lambda_{i,k,j}^1=0, \lambda_{i,k,j}^2=0, \ldots, \lambda_{i,k,j}^Q=0$\\
 \For{$x \in S_i$} {
	
	$q = \max \{1, \lceil (v_{k,j}(x)-v^\text{min}_{i,k,j})/\Delta_{i, k, j} \rceil \}$\\
	$\lambda_{i,k,j}^q$++

 }
 $\lambda_{i,k,j}^1/=|S_i|, \lambda_{i,k,j}^2/=|S_i|, \ldots, \lambda_{i,k,j}^Q/=|S_i|$\\ 
 
 $\sigma_{i, k, j} = (v_{i,k,j}^\text{min}, v_{i,k,j}^\text{max}, \lambda_{i,k,j}^1, \lambda_{i,k,j}^2, \ldots, \lambda_{i,k,j}^Q)$  \\
 Add $\sigma_{i, k, j}$ to $\sigma_i$ \\
 }	
 }
 \Return $\sigma = \{\sigma_1, \ldots, \sigma_{m} \}$ \\
\caption{Aggregation Algorithm of MRC.}
\label{alg:agg-MRC}
\end{algorithm}

The idea behind the online phase of MRC is that the more a new input $x_{new}$ produces neuron output values outside the sub-ranges matched by the inputs in the Trusted Set, the more $x_{new}$ is likely to be unsafe.
Given $x_{\text{new}}$, the DNN Active State is composed of the identifiers of the sub-ranges to which the output 
of the neurons belong to, if any.
This is used to assign a cost $\Theta_{k,j}(x_{\text{new}}) \in [0, 1]$ to each neuron $n_{k,j}$ that quantifies how much the neuron output is deemed unsafe. Being $i$ the index of the class assigned to $x_{\text{new}}$, this cost is formally defined as
\begin{equation}
\Theta_{k,j}(x_{\text{new}}) = 
\begin{cases} 
1	, & \mbox{if }v_{n_{k,j}}(x_{\text{new}}) \notin [v^\text{min}_{i,k,j}, v^\text{max}_{i,k,j}] \\ 
1-\lambda_{i,k,j}^{q^*}, & \mbox{otherwise,}
\end{cases}
\end{equation}
where $q^*$ denotes the index of the sub-range of $n_{k,j}$ to which $v_{n_{k,j}}(x_{\text{new}})$ belongs to.

The Confidence Evaluation Algorithm, reported in Algorithm~\ref{alg:safety-eval-MRC}, computes the sum of such costs $\Theta_{k,j}(x_{\text{new}})$ over all neurons, denoted here as $\eta$. 
Hence, the confidence score is computed as $c = \exp({- \frac{\eta \cdot ln(2)}{\tau_{\hat{y}}}})$.

\begin{algorithm}[htb!]
\SetAlgoLined
 \textbf{Input} input $x_{new}$, DNN Signature $\sigma$, trained DNN, thresholds $\tau_i$, number of sections $Q$ \\
 \textbf{Output} confidence $c$ \\
 \algrule
 $\hat{y} = \text{argmax}_{1\leq j \leq m} \{ f(x_{\text{new}})\}$\\
 $\eta$ = $0$ \\
 \For{$n_{k,j} \in N$} {
 Extract $\sigma_{\hat{y}, k, j}$ from $\sigma_{i}$\\
 Extract $v^\text{min}_{\hat{y},k,j}$ and $v^\text{max}_{\hat{y},k,j}$ from $\sigma_{\hat{y}, k, j}$\\
 \If{$v_{k,j}(x_{\text{new}}) \notin [v^\text{min}_{\hat{y},k,j}, v^\text{max}_{\hat{y},k,j}] $}  {
 $\eta$ += 1
 }
 \Else{
 	$\Delta_{\hat{y}, k, j} = (v_{\hat{y},k,j}^\text{max}-v_{\hat{y},k,j}^\text{min})/Q$\\
 	$q^* = \max \{1, \lceil (v_{k,j}(x_{\text{new}})-v^\text{min}_{\hat{y},k,j})/\Delta_{\hat{y}, k, j} \rceil \}$\\
 	Extract $\lambda_{\hat{y},k,j}^{q^*}$ from $\sigma_{\hat{y}, k, j}$\\
 	$\eta$ += $1-\lambda_{\hat{y},k,j}^{q^*}$ \\
 }
 }
 
 \Return $c = \exp({- \frac{\eta \cdot ln(2)}{\tau_{\hat{y}}}})$ \\

 \caption{Confidence Evaluation Algorithm of MRC.}
 \label{alg:safety-eval-MRC}
\end{algorithm}

\subsection{k-Nearest Neighbors Coverage (kNNC)}

This CAM is based on the DkNN algorithm presented in~\cite{k_nearest}, which was originally proposed 
as an alternative inference logic and not as a way to detect unsafe inputs.
The key idea of the kNNC is to compute a confidence score by applying the k-Nearest Neighbors (kNN) algorithm on the 
output values produced by the neurons of the various DNN layers.

For each output class with index $i$, the DNN Signature is 
composed of a collections of sets $\sigma_{i,k}$ of vectors, one for each layer $L_k$.
Each set $\sigma_{i,k}$ is composed by aggregating the vectors $V_{k}(x)$ obtained from all inputs $x \in S_i$.
The corresponding procedure is reported in Algorithm~\ref{alg:agg-kNNC}.

\begin{algorithm}[htb!]
\SetAlgoLined
 \textbf{Input} Trusted Set $S$, trained DNN \\
 \textbf{Output} DNN Signature $\sigma$ \\
 \algrule
\For{$S_{i} \in S$} { 
  $\forall k=1, \ldots, K, ~ \sigma_{i, k} = \{ \}$\\
 \For{$x \in S_i$}{
 \For{$L_k \in L$}{
 Add $V_{k}(x)$ to $\sigma_{i,k}$
 }
 }
 $\sigma_i = \{ \sigma_{i, 1}, \ldots, \sigma_{i, K} \}$
}
\Return $\sigma = \{\sigma_1, \ldots, \sigma_{K}\}$ \\
\caption{Aggregation Algorithm of kNNC.}
\label{alg:agg-kNNC}
\end{algorithm}

During the online phase of kNNC, given a new input $x_{\text{new}}$ and its corresponding predicted class $\hat{y}$, the kNN algorithm is used to compute the confidence value.
In the following, the parameter that controls the kNN algorithm is denoted as $G$, since $k$ denotes the layer index. 
Hence, the $G$ vectors $V_{k}(x)$ stored in the DNN Signature that are nearest to $V_{k}(x_{\text{new}})$ are identified by the kNN algorithm and their corresponding output class indexes are recorded and stored in a multiset $\Omega_k$.
Finally, the number of occurrences of the class index $\hat{y}$ in $\Omega_k$ is counted and denoted as $\eta$. Hence, the output confidence score is computed as $c = exp(-\frac{\eta \cdot ln(2)}{\tau_{\hat{y}}})$.
The rationale behind this computation is the following. Given inputs $x$ classified to the $\hat{y}$-th class by the DNN, the more vectors $V_{k}(x)$ in the signature are among the $G$ closest ones to $V_{k}(x_{\text{new}})$, the more the active state produced by $x_{\text{new}}$ resembles the ones produced by the trusted inputs that the DNN classifies as for $x_{\text{new}}$, hence positively contributing to the confidence value.
The overall procedure is reported in Algorithm~\ref{alg:safety-eval-kNNC}.

\begin{algorithm}[htb!]
\SetAlgoLined
 \textbf{Input} input $x_{new}$, DNN Signature $\sigma$, trained DNN, thresholds $\tau_i$, number of nearest neighbors $G$ \\
 \textbf{Output} confidence $c$ \\
 \algrule
 $\hat{y} = \text{argmax}_{1\leq j \leq m} \{ f(x_{\text{new}})\}$\\
 $\eta$ = $0$ \\
 \For{$L_k \in L$}{
 	$\Omega_{k} = \text{kNN}(G, V_{k}(x_{\text{new}}),\sigma_{1,k}, \ldots, \sigma_{m, k})$\\
 	 $\eta$ +=  $ |{y \in \Omega_{k} : y = \hat{y}}| $\\
 }
 \Return $c = exp(-\frac{\eta \cdot ln(2)}{\tau_{\hat{y}}})$ \\

\caption{Confidence Evaluation of kNNC.}
\label{alg:safety-eval-kNNC}
\end{algorithm}

\subsection{Illustrative example}

To illustrate the proposed approach, let us consider a simple example based on an inspired LeNet-4~\cite{lecun} CNN trained on the MNIST dataset~\cite{mnist}.
The Trusted Set consists of those MNIST samples that are correctly predicted by the network with a prediction score higher than $0.9$. 
The selected CAM is the SRC presented in Section~\ref{s:SRC}.
Unsafe inputs are generated using the FGSM~\cite{goodfellow2014explaining}, setting $\epsilon=0.05$ to obtain AEs with small perturbations.

\begin{figure}[tp]
	\centering
    \subfloat[\label{fig:detectionExA}]{\includegraphics[scale=0.40]{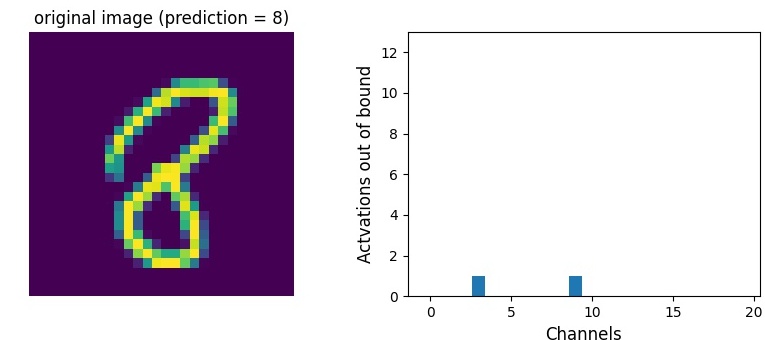}
    }
    \hfill

    \subfloat[\label{fig:detectionExB}]{\includegraphics[scale=0.40]{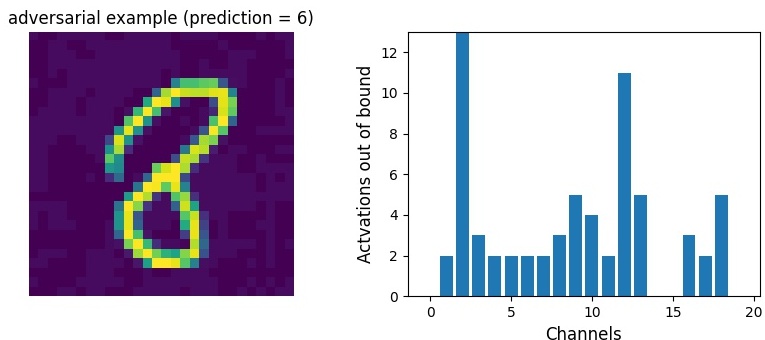}
    }
    \caption{(a) A genuine sample and its corresponding activation histogram with respect to a DNN Signature obtained in the first convolutional block of LeNet, under the SRC. (b) An adversarial example and its corresponding activation histogram against the same DNN Signature.}

\end{figure}

For simplicity, the SRC method is only applied to the first convolutional block, including a convolutional layer, a pooling layer, and a ReLU activation layer.
Figure~\ref{fig:detectionExA} illustrates a trusted input (i.e., a digit correctly classified as `8' with a prediction score of $0.974$), which is not part of the Trusted Set, together with a histogram that reports the number of neurons per channel that have an output value outside the corresponding ranges $[v^\text{min}_{i,k,j}, v^\text{max}_{i,k,j}]$ representing the DNN Signature for class `8' under the SRC. In particular, note that only two neurons have outputs that do not match the signature, producing a prediction confidence $c = \exp({- \frac{ \mathbf{2} \cdot ln(2)}{\tau_8}}) = 0.8705$, where in this example all the thresholds $\tau_i$ are set to 10. 
%
%
Conversely, Figure~\ref{fig:detectionExB} illustrates an adversarial example, incorrectly classified as `6' (with a prediction score of $0.918$), together with the corresponding histogram. In this case, there are 63 neurons in the first block whose activations are outside the ranges stored in the DNN Signature for label `6', resulting in a prediction confidence $c = \exp({- \frac{ \mathbf{63} \cdot ln(2)}{\tau_6}}) = 0.0126$.

\section{Implementation in Caffe} \label{s:impl}

\begin{figure}[tp]
    \centering
    \subfloat[\label{fig:caffe-impl}]{\includegraphics[width=.95\linewidth]{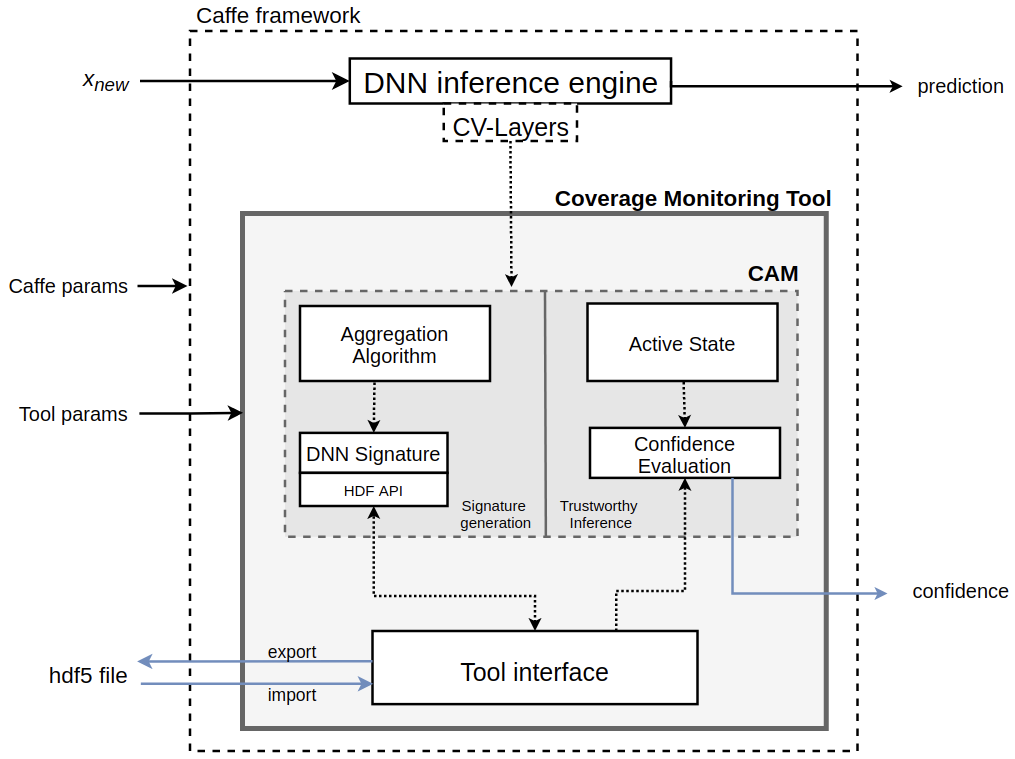}}
    \hfill

    \subfloat[\label{fig:cv-alyer-impl}]{\includegraphics[width=.85\linewidth]{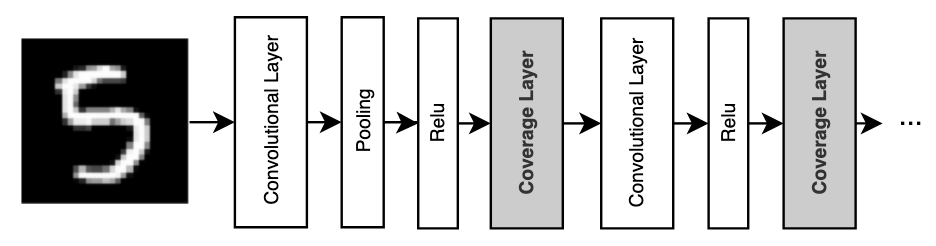}}
    \caption{(a) Overview of the tool architecture. (b) Example network architecture with two CV-Layers installed after two ReLu activation layers.}
    \vspace{-1em}
\end{figure}

The proposed architecture has been implemented as an extension of the Caffe framework ~\cite{caffe} --- see Figure~\ref{fig:caffe-impl}.
The extension introduces a new type of layer in Caffe, called \emph{Coverage Layer} (CV-Layer), which operates in a transparent (i.e., pass-through) fashion, hence not altering neither the DNN performance nor the DNN hyperparameters.
The CV-Layer applies a coverage criterion at inference time based on the tensor data it receives in input and is used during both the offline and online phases. When installing a CV-Layer, given that it operates as a pass-through component, 
connections between two layers, say $L_k$ and $L_{k+1}$, can be preserved by connecting (i) the output of $L_k$ to the input of the CV-Layer and (ii) the output of the CV-Layer to the input of $L_{k+1}$.
Multiple CV-Layers can be installed depending on the behavior of the selected CAM. For instance, CAMs that work by analyzing all neuron outputs of the network require the installation of a CV-Layer after each layer of the original DNN.
An example installation of the CV-Layer is illustrated in Figure~\ref{fig:cv-alyer-impl}.
 


%

Figure~\ref{fig:caffe-impl} illustrates the interconnection between the principal software blocks of the tool and the Caffe framework.
Under Caffe, networks are distributed via a \texttt{prototxt} model file, which should include the CV-layers in the network architecture, and a \texttt{caffemodel} file, which contains the trained weights (both referred to as 'Caffe params' in the figure).
Regarding the tool, the required parameters (referred to as 'Tool params' in the figure) are a specification of the Trusted Set and the identifier of the selected CAM (to be chosen among those supported by the tool).
The tool will then instantiate the implementation of the monitoring functions associated to the selected CAM.
Furthermore, CAM-specific parameters can be specified (see Section~\ref{s:CAMs}).
Finally, the tool is capable of exporting and importing the DNN Signature using the hdf5 file format by means of the hierarchical data format (HDF) API.
The tool provides an interface to invoke monitoring operations. 
It manages the interaction with the DNN Signature, as the export at the end of the offline phase, or the import at the beginning of the online phase.
The tool also allows reading the confidence evaluation outcome during the online phase, which represents the coverage metric of the applied CAM.

Each CAM is developed both in CUDA and C++. Since the CV-Layers are totally compliant with the other Caffe layers, the DNN framework will take care of forwarding tensors to the selected implementation based on the available architecture (e.g., GPU or CPU). 
At the same time, it is up to the tool to select the device implementation of the Confidence Evaluation Algorithms, which are performed at the end of the DNN inference. 

Another feature offered by the tool is the possibility of computing the signature in the offline phase using a mini-batch approach. 
The Trusted Set is split into multiple mini-batches where each one represents a single, but large input tensor that is analyzed by the CV-Layer on the same inference pass. 
The values extracted from the inputs of the mini-batch, according to the selected CAM, are then merged into the DNN Signature with the results obtained by previously-processed mini-batches.
This approach significantly allows speeding up the creation of the DNN Signature creation time, 
especially when it is possible to parallelize the procedure.

\section{Experimental evaluation} \label{s:exps}
{\CHANGE
This section reports the results of a set of experiments conducted 
to assess the performance of the proposed methods in terms of running time, memory footprint, and capability of detecting unsafe network inputs.
Several types of unsafe inputs generated with state-of-the-art methods have been considered: to make the paper self-contained, they are summarized in Section~\ref{s:unsafe_inputs} and described in details in the supplementary material. 
Before discussing the actual experimental results, Section~\ref{s:exp_setting} presents the experimental setting, whiles Section~\ref{s:exp_calib} presents the calibration of thresholds used.

\subsection{Unsafe inputs generation} \label{s:unsafe_inputs}

In recent years, many adversarial attacks~\cite{szegedy2014intriguing} have been disclosed to efficiently generate AEs while minimizing the perturbation to be applied to turn a safe input into an unsafe one. Besides AEs, a network can also fail for other types of unsafe inputs.
A  recap of all the methods to generate unsafe inputs tested during our evaluation is reported next, while their complete description is provided in the supplementary material accompanied by illustrative examples. 

\textbf{Adversarial examples.}
We considered the following attack methods under different settings for crafting both small and medium perturbations. The amount of perturbation is usually refereed with the $\epsilon$ parameter. 
Among the first and famous adversarial attacks, one is FGSM~\cite{goodfellow2014explaining}, which crafts adversarial perturbation in one shot.
More advanced adversarial attacks have also been considered: they are PGD~\cite{pgd_attack}, BIM~\cite{BIM} and CW~\cite{CarliniAttack}, which implement iterative methods capable of improving the attack effectiveness at each iteration. 
The number of iterations and the amount of perturbation introduced at each iteration are denoted by $k$ and $\alpha$, respectively.
The attacks are configured as reported in Table~ \ref{table:set_test}.

\textbf{Out-of-distribution unsafe inputs.}
Another class of unsafe inputs are those that are far from the distribution of data used to train the network, but are still predicted with a high score by the model. 
They commonly exist because a DNN works as a global classifier for the whole input space (e.g., all possible images of a given size and format).
To test this class of unsafe inputs, we generated new samples with large perturbations, obtained with a variant of the FGSM attack, which are applied to samples in the training set of the tested networks. Details on the generations are reported in the supplementary material.

\textbf{Adversarial patches.}
Finally, we also studied another famous class of adversarial images, namely those using patches to make them adversarial. 
Adversarial patches are crafted with the PGD~\cite{pgd_attack} method, but perturbing a limited area of the image only.
}

\subsection{Experimental setting} \label{s:exp_setting}

The datasets considered in the experimental evaluation are MNIST~\cite{mnist} and F-MNIST~\cite{fmnist}.
MNIST is a dataset of handwritten digits while F-MNIST is a dataset of clothes articles. Both datasets contain grayscale images of 28x28 pixels, and provide 60,000 images for training and 10,000 images for testing.  
F-MNIST is slightly more complex than MNIST and DNNs usually achieve lower accuracy in classifying its images.
Both the datasets have been processed with a LeNet-4~\cite{lecun} CNN trained on 8 epochs using the Adam optimization algorithm~\cite{adam} and a cross entropy loss function.

{\NEWCHANGE It is of utmost importance to note that all the related works close to this paper, i.e., those that adopt internal network analysis with procedural algorithms only (discussed in Section~\ref{s:relatedwork}), 
performed experimental evaluations on data sets and DNNs with similar complexity to the ones mentioned above.}
This is because the research on methods for increasing the confidence of DNNs is still in an early stage and, as such, the proposals have to be first evaluated on small-size network models.


Three Coverage Layers per network have been installed (other network models with different installations of Coverage Layers are discussed later).
Table\ref{table:model_tab} summarizes the network architectures and their corresponding classification accuracy on the original testing sets.

\begin{table}[!t]
\centering
\resizebox{0.5\textwidth}{!}{%
\centering
\begin{tabular}{cccll}
\cline{1-3}
\multicolumn{1}{|c|}{Dataset} & \multicolumn{1}{c|}{Model}                                                                                           & \multicolumn{1}{c|}{Accuracy} &  &  \\ \cline{1-3}
\multicolumn{1}{|c|}{MNIST}   & \multicolumn{1}{c|}{\makecell{conv(20,5,1) - ReLU - \textbf{CV} - MaxPool(2,2) - conv(50,5,1)\\ - ReLU - \textbf{CV} - fc(500) - ReLU - \textbf{CV} - fc(10)}} & \multicolumn{1}{c|}{0.9907}   &  &  \\ \cline{1-3}
\multicolumn{1}{|c|}{F-MNIST} & \multicolumn{1}{c|}{\makecell{conv(20,5,1) - ReLU - \textbf{CV} - MaxPool(2,2) - conv(50,5,1)\\ - ReLU - \textbf{CV} - fc(500) - ReLU - \textbf{CV} - fc(10)}} & \multicolumn{1}{c|}{0.9106}   &  &  \\ \cline{1-3}
\multicolumn{1}{l}{}          & \multicolumn{1}{l}{}                                                                                                 & \multicolumn{1}{l}{}          &  & 
\end{tabular}
}
\vspace{-1.5em}
\caption{\small{Network models applied for the experimental evaluation. 'CV' indicates a Coverage Layer.}}
\vspace{-1.5em}
\label{table:model_tab}
\end{table}

The experimental evaluation used the following sets of inputs:

\begin{itemize}
\item \textit{Trusted Set}: it is the one used to generate the DNN Signature and has been obtained by selecting from the original \emph{training set} those samples that the DNN classifies correctly with a prediction score (i.e., the softmax probability) larger than $0.9$.
This set contains 59309 and 52680 samples for MNIST and F-MNIST, respectively.

\item \textit{Trusted Test Set}: it is the one used to assess the classification performance of the DNNs when enhanced with
one of the proposed CAMs. It is obtained by selecting from the original \emph{testing set} those samples that the DNN classifies correctly with a prediction score $> 0.9$.
This set contains 9000 samples.

\item \textit{Adversarial Set}: it is the one employed for evaluating the performance of the proposed methods in detecting unsafe inputs. It contains unsafe inputs for which the DNN makes a wrong prediction.
Different definitions of this set have been tested depending on the selected attack method (see Section~\ref{s:unsafe_inputs}): they are summarized in Table~\ref{table:set_test} for both MNIST and F-MNIST.
{\CHANGE A prediction is considered to be wrong when the DNN classifies such inputs in a wrong class with a prediction score larger than $0.8$, for AEs and adversarial patches, or larger than $0.99$, for out-of-distribution inputs.
Such thresholds have been selected by empirically finding the largest values that allowed generating a sufficient number of unsafe inputs to perform the experiments. Note that the higher the thresholds the more difficult the generation of AEs. Conversely, when using lower thresholds, the generated AEs tend to be less relevant, as they could be simply discarded by comparing the softmax score produced by the network against a certain threshold. Further details are provided in the supplementary material.
}
\end{itemize}

\begin{table*}[!t]
\resizebox{\linewidth}{!}{%
\begin{tabular}{|c|c|c|c|c|c|c|}
\hline
              & \multicolumn{3}{c|}{\textbf{MNIST}}                                                       & \multicolumn{3}{c|}{\textbf{F-MNIST}}                                                     \\ \hline
Attack method & Attack parameters       &  \makecell{\# of samples for \\calibration} &  \makecell{\# of samples for \\evaluation}  & Attack parameters       &  \makecell{\# of samples for \\calibration}  &  \makecell{\# of samples for \\evaluation} \\ \hline
FGSM-1        & $\epsilon = 0.1  $                                               & 500                                                                                       & 500                                                                                      & $\epsilon = 0.05   $                                             & 600                                                                                       & 3400                                                                                     \\ \hline
FGSM-2        & $\epsilon = 0.2  $                                               & 800                                                                                       & 3200                                                                                     & $\epsilon = 0.1   $                                              & 600                                                                                       & 1400                                                                                     \\ \hline

PGD-1        & $(\epsilon, \alpha, k) = (0.1, 0.015, 40)$                       & 800                                                                                       & 1800                                                                                     & $(\epsilon, \alpha, k) = (0.03, 0.015, 40)$                      & 600                                                                                       & 3400                                                                                     \\ \hline 
PGD-2         & \multicolumn{1}{l|}{$(\epsilon, \alpha, k) = (0.18, 0.015, 40)$} & 800                                                                                       & 3200                                                                                     & \multicolumn{1}{l|}{$(\epsilon, \alpha, k) = (0.10, 0.015, 40)$} & 600                                                                                       & 3400                                                                                     \\
\hline
BIM-1           & $(\epsilon, \alpha, k) = (0.05, 0.004, 10)$                      & 450                                                                                       & 500                                                                                      & $(\epsilon, \alpha, k) = (0.015, 0.004, 10)$                     & 600                                                                                       & 1500                                                                                     \\ \hline
BIM-2           & $(\epsilon, \alpha, k) = (0.18, 0.004, 10)$                      & 800                                                                                       & 3200                                                                                     & $(\epsilon, \alpha, k) = (0.10, 0.004, 10)$                      & 600                                                                                       & 3400                                                                                     \\ \hline
CW            & $k=500$                                  & 450                                                                                       & 500                                                                                      & $k=500$                                  & 200                                                                                       & 300                                                                                      \\ \hline
Out of Dis.   & $\epsilon =0.02, k=80  $                                         & -                                                                                         & 4000                                                                                     & $\epsilon =0.02, k=80 $                                          & -                                                                                         & 4000                                                                                     \\ \hline
Patch         & $\epsilon = 0.02, k = 200$                                       & -                                                                                         & 100                                                                                      & $\epsilon = 0.02, k = 200$                                       & -                                                                                         & 1000                                                                                     \\ \hline
\end{tabular}
}
\vspace{-0.8em}
\caption{\small{Parameters and settings used to generate the Adversarial Set.}}
\vspace{-0.8em}
\label{table:set_test}
\end{table*}

%

%

{\CHANGE
In the following, the experimental results are reported for four CAMs: SRC, two versions of MRC with $Q = \{ 16, 32\}$, and kNNC with $G = 75$.
Furthermore, we also compared the results against state-of-the-art run-time methods used for detecting AEs: FeaturesSqueezing \cite{feature_squeezing} and VisionGuard \cite{vision_guard}. Details on their behavior and implementation are provided in the supplementary material. 
}

\subsection{Threshold calibration} \label{s:exp_calib}
{\CHANGE
This experimental evaluation was focused on a binary classification of the network inputs, i.e., either an input was deemed \emph{safe} 
and the prediction made by the network was accepted, or the input was deemed \emph{unsafe}, and the network prediction was rejected.
This has been implemented by calibrating the tolerances $\tau_i$ used by the various Confidence Evaluation Algorithms so that 
an input is deemed safe if $c \geq 0.5$ and unsafe otherwise.

The calibration of thresholds has been performed using Receiver Operating Characteristic (ROC) analysis to compute 
the values $\tau_i$, for each class with index $i=1, \ldots, m$, that represent the best balance between 
minimizing the inputs that are wrongly rejected and accepted.
In this regard, portions of the Trusted Test Set and the Adversarial Set introduced in the previous section have been used to test safe and unsafe inputs, respectively, during calibration.
A more detailed discussion of the calibration phase and an illustration of the obtained ROC curves is presented in the supplementary material. 
}

\subsection{Detection performance} \label{s:exp_detection}

Experiments have been conducted to evaluate the performance of the four CAMs in correctly detecting whether an input is safe or unsafe using the thresholds $\tau_i$ calibrated as described above.
Tables~\ref{table:test} reports the detection accuracy, defined as the ratio of inputs \emph{correctly} classified as either safe or unsafe. The tables report the breakdown of unsafe inputs for all the methods introduced in Section~\ref{s:unsafe_inputs} (see Table~\ref{table:set_test} for the number of samples used for the evaluation).
The last rows of the tables are related to safe inputs, which correspond to the remaining inputs of the Trusted Test Set that have not been considered for calibration.

\begin{table*}[!t]

\centering
\resizebox{\linewidth}{!}{%
\begin{tabular}{|c|c|c|c|c|c|c|c|c|c|c|c|c|}
\hline
             & \multicolumn{6}{c|}{MNIST}                                                                   & \multicolumn{6}{c|}{F-MNIST}                                                                 \\ \hline
Input type   & \textit{SRC} & \textit{MRC-32} & \textit{MRC-16} & \textit{kNNC} & \textit{VG} & \textit{FS} & \textit{SRC} & \textit{MRC-32} & \textit{MRC-16} & \textit{kNNC} & \textit{VG} & \textit{FS} \\ \hline
FGSM-1       & 0.998                         & 0.976                            & 0.946                            & 0.958                          & 0.986                        & 0.976                        & 0.855                         & 0.848                            & 0.827                            & 0.915                          & 0.880                        & 0.881                        \\ \hline
FGSM-2       & 1.0                           & 0.994                            & 0.989                            & 0.941                          & 0.930                        & 0.936                        & 0.929                         & 0.952                            & 0.937                            & 0.883                          & 0.785                        & 0.814                        \\ \hline
PGD-1          & 0.989                         & 0.968                            & 0.943                            & 0.970                          & 0.989                        & 0.976                        & 0.852                         & 0.841                            & 0.792                            & 0.935                          & 0.901                        & 0.973                        \\ \hline
PGD-2          & 1.0                           & 0.919                            & 0.852                            & 0.955                          & 0.810                        & 0.882                        & 0.979                         & 0.942                            & 0.943                            & 0.859                          & 0.175                        & 0.325                        \\ \hline
BIM-1          & 0.968                         & 0.943                            & 0.912                            & 0.964                          & 1.0                          & 0.985                        & 0.842                         & 0.848                            & 0.789                            & 0.916                          & 0.997                        & 1.0                          \\ \hline
BIM-2          & 0.995                         & 0.861                            & 0.815                            & 0.952                          & 0.8425                       & 0.870                        & 0.986                         & 0.924                            & 0.939                            & 0.831                          & 0.161                        & 0.239                        \\ \hline
CW           & 0.971                         & 0.959                            & 0.933                            & 0.960                          & 1.0                          & 0.989                        & 0.813                         & 0.829                            & 0.822                            & 0.907                          & 0.993                        & 1.0                          \\ \hline
Out of Dis.  & 1.0                           & 1.0                              & 1.0                              & 0.346                          & 0.0                          & 0.0                          & 1.0                           & 1.0                              & 1.0                              & 0.842                          & 0.00                         & 0.00                         \\ \hline
Patch          & 1.0                           & 0.891                            & 0.923                            & 0.995                          & 0.957                        & 0.984                        & 1.0                           & 0.911                            & 0.858                            & 0.951                          & 0.962                        & 0.975                        \\ \hline
Safe Samples & 0.991                         & 0.951                            & 0.919                            & 0.955                          & 0.925                        & 0.938                        & 0.89                          & 0.892                            & 0.851                            & 0.921                          & 0.671                        & 0.732                        \\ \hline
\end{tabular}
}
\vspace{-0.8em}
\caption{\small{Detection accuracy of all the tested methods on MNIST and F-MNIST. {\CHANGE VG and FS correspond to Vision Guard \cite{vision_guard} and FeatureSqueezing \cite{feature_squeezing}, respectively.}
}}
\label{table:test}
\vspace{-0.8em}
\end{table*}

{\CHANGE
Although SRC is the simplest CAM, for inputs of the MNIST dataset it is often capable of detecting unsafe inputs better than the others CAMs.
The only exception pertains to the BIM attack, but the performance of SRC is anyway extremely close to the one of kNNC.
Note that SRC and MRC are also capable of detecting all unsafe out-of-distribution inputs, while kNNC exhibits very poor performance 
(close to coin tossing) for that kind of inputs.

The results for the F-MNIST dataset are quite different. In this case, there is no dominating CAM for AEs.
Note that kNNC performs quite well with AEs generated using low perturbations (e.g., CW and BIM), while SRC significantly outperforms 
kNNC for unsafe inputs generated using the PGD attack. 
SRC and MRC are again capable of detecting all unsafe out-of-distribution inputs.
This suggests that, given proper accelerated implementations, multiple CAMs could be used in parallel to improve the overall detection performance of unsafe inputs.

Table \ref{table:test} also provides the results achieved with Vision Guard and FeatureSqueezing. They always obtain high performance with low-perturbation attacks on both MNIST and FMNIST, but fail when the magnitude of the perturbation increases (e.g., PGD-2 and BIM-2, which find more effective AEs). Moreover, they report worse performance on safe samples, since they are incorrectly rejected more times than the proposed CAMs. Most interestingly, Vision Guard and FeatureSqueezing totally fail with out-of-distributions unsafe inputs, while three of the proposed CAMs are very effective.

Finally, to further evaluate the detection performance of the CAMs, additional tests are provided in the supplementary materials to better evaluate the relationship between the CAMs and the perturbations of AEs. 
}

\subsection{Running time and memory footprint} \label{s:cost_perf}
With respect to a regular deployment of a DNN, each of the proposed CAMs introduces an additional inference time to execute 
the Confidence Evaluation Algorithm (implemented by CV-Layers) and an additional memory footprint to store the DNN Signature.
This section is focused on evaluating these overheads, which in some cases may be very important in selecting the most appropriate CAM.
For instance, to deploy a CAM on a resource-constrained, embedded device that has to operate in real time, it is essential 
to contain as much as possible the additional inference time and the memory footprint, even accepting reduced detection performance.

\begin{figure}[!t]
\centering
\makebox[\columnwidth]{\includegraphics[scale=0.30]{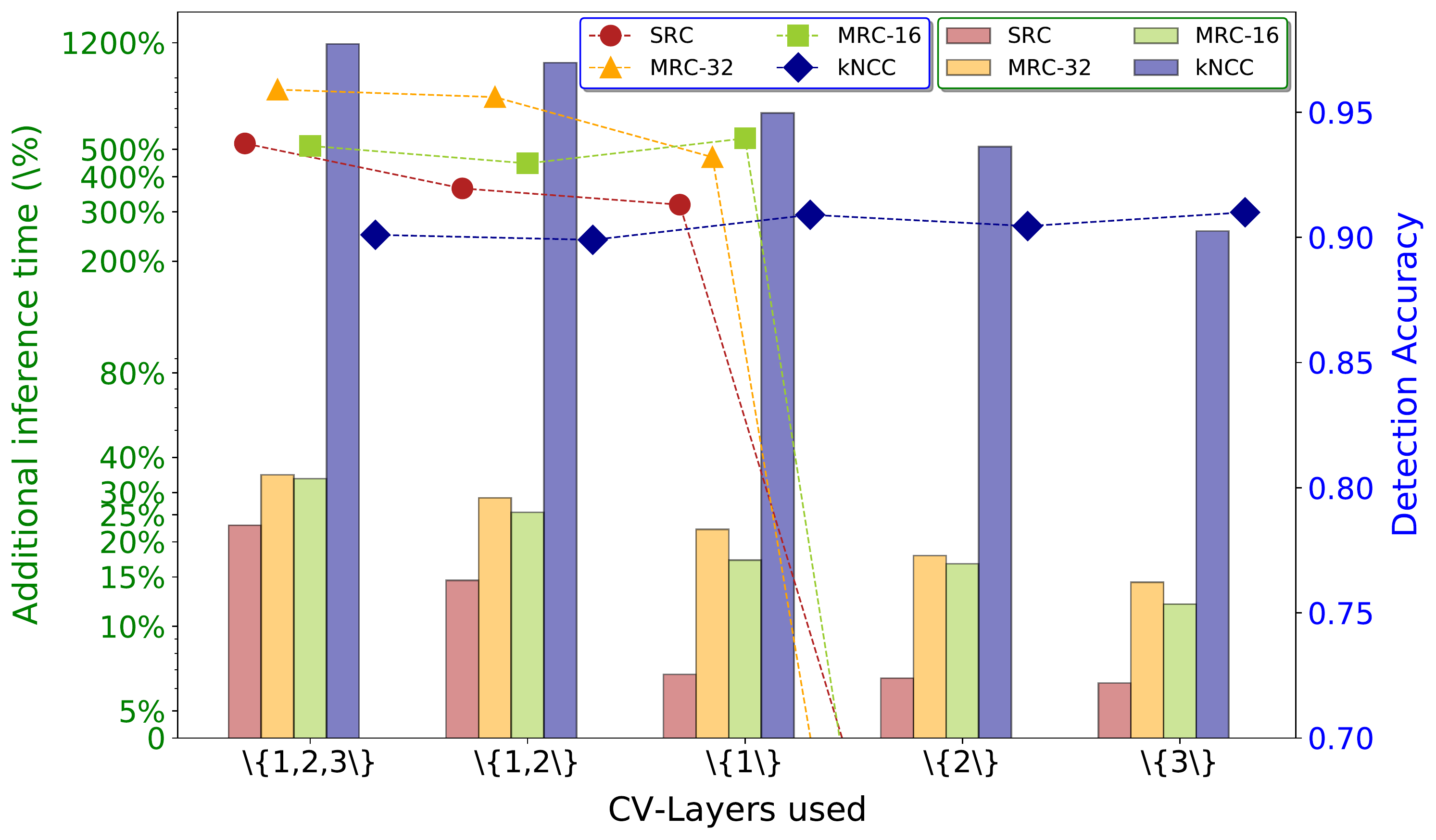}}
\vspace{-2.0em}
\caption{\small{Timing performance and detection accuracy for five installations of the CV-Layers. The sets on the x-axis denote which of the CV-Layers 
reported in Table~\ref{table:model_tab} are installed (numbered in the order with which they appear in the table). For instance, $\{1, 2\}$ refers to the case in which only the first two CV-Layers are installed.}}
\vspace{-1.2em}
\label{fig:timing_perf}
\end{figure}

\textbf{Additional inference time.}
The experiments for SRC, MRC, and NRC have been performed on a machine equipped with an Intel(R) Core(TM) i5-4670K CPU @ 3.40GHz, 8GB of RAM, and an NVIDIA GeForce GTX770 GPU. Due to their demanding memory requirements (see the results at the end of this section), the experiments for kNNC have been performed on a Nvidia DGX Station V100.
The implementation presented in Section~\ref{s:impl} has been used\footnote{Note that, due to implementation issues under resolution in integrating kNNC in our CUDA-based implementations of CV-Layers, the experiments for kNNC have been performed with a Python implementation of the tool presented in Section~\ref{s:impl} that uses the Pytorch framework~\cite{pytorch} (with GPU acceleration) and FAISS~\cite{faiss}, which is notably the best-performing library that implements a GPU-accelerated exact nearest neighbors search.}.
Figure~\ref{fig:timing_perf} reports the additional inference time introduced by the four CAMs
together with the corresponding detection performance for a representative setting (F-MNIST dataset and unsafe inputs generated by the FGSM-2 attack). 
Note that the y-axis at the left of the figure has a logarithmic scale.
The figure reports the results for different installations of the CV-Layers, denoted by the sets reported on the x-axis of the plot.
They correspond to cases in which the CAMs operate on a subset of the entire set of CV-Layers reported in Table~\ref{table:model_tab}.
The calibration procedure presented in Section~\ref{s:exp_calib} has been performed for each of them (i.e., each installation is assigned 
different thresholds).

As it can be noted from the figure, when all the 3 CV-Layers are installed ($\{$1,2,3$\}$ on the x-axis) 
the percentage increase of the interference time (with respect the original network inference) is of the same order of magnitude for SRC, MRC-16, and MRC-32. As one may easily expect, the best timing performance is achieved by SRC, with a percentage increase of 22.9$\%$.
The additional inference time introduced by kNNC is instead quite high and corresponds to a 1188$\%$ increase.

Interesting observations can be made by looking at the accuracy of the various installations and CAMs: SRC and MRC exhibit very 
poor detection performance if the first CV-Layer is not installed, while kNNC exhibits only slights variations of the detection performance 
when less CV-Layers are installed. Most interestingly, note that the detection performance may even increase when less CV-Layers are installed.
For instance, these results reveal that it is not convenient to adopt kNNC with the first two CV-Layers installed, as the same detection 
performance can be achieved with just a 255$\%$ increase of the inference time, rather than a 1188$\%$ increase for the case in which all CV-Layers are installed.



\textbf{Memory footprint.}
Table \ref{table:memory_table} reports the memory footprint of the DNN Signature of the various CAMs for each CV-Layer.
As it can be noted from the table, SRC and NRC have a very modicum memory footprint (in the order of 1MB in total), while 
kNNC is characterized by a huge memory footprint that even exceeds 3GB if all the three CV-Layers are installed.
These results, together with the ones of Figure~\ref{fig:timing_perf}, confirm that under kNNC it does not worth to install CV-Layers 
between shallower layers of the network, and that SRC is a very good choice to balance performance with overheads. The results for the kNNC may vary if different (e.g., approximate) methods to implement the nearest neighbors search are adopted.

{\CHANGE

\textbf{Comparison with other detection techniques.} The average running times of both VG and FS were also measured. They introduce an additional latency of $138\%$ and $237\%$, respectively, which correspond to the running time of multiple inferences, plus an additional overhead introduced by the used transformations. 
As far memory footprint is concerned, they are very lightweight approaches only when the multiple inferences they require are sequentially performed (as in the evaluated setting). Conversely, if they are performed in parallel, the overall memory footprint grows as a function of the amount of space needed for instantiating and running multiple replicas of the network model.
}

\begin{table}[!t]
\resizebox{\linewidth}{!}{%
\centering
\begin{tabular}{|c|c|c|c|}
\hline
\multicolumn{1}{|l|}{} & CV-Layer1 & CV-Layer2 & CV-Layer3 \\ \hline
SRC                    & 944 kB    & 278 kB    & 62,8 kB    \\ \hline
MRC-16                 & 8,3 MB    & 2,3 MB    & 387 kB     \\ \hline
MRC-32                 & 15,7 MB   & 4,4 MB    & 707 kB     \\ \hline
NRC                    & 483,6 kB  & 150,8 kB  & 42,8 kB   \\ \hline
kNNC                   & 2.6 GB    & 732 MB    & 114 MB    \\ \hline
\end{tabular}
}
\caption{Memory footprint of the DNN Signature (exported HDF file) of the various CAMs for each CV-Layer.}
\label{table:memory_table}
\end{table}

\subsection{Adaptive attack and countermeasure} \label{s:full_kn}
{\NEWCHANGE
Although the presented CAMs are able to detect several kinds of unsafe inputs, ad-hoc attacks can still be devised to optimize unsafe inputs by 
directly exploiting the knowledge of the detection mechanism and the signatures.

Inspired by the famous gradient-based attacks (which are formally discussed in the supplementary material), we designed a white-box attack method, called \textit{Signature-Attack}, for crafting adversarial inputs while trying to keep the activation of neurons within the range of the signatures.
This is accomplished by optimizing the adversarial perturbation through two loss functions, $\mathcal{L}_{CE}$ and $\mathcal{L}_{S}$. 
The former is the common cross-entropy loss, also involved in all the other attacks tested in this work. The optimization process shall maximize such a loss to intensify the adversarial effect of the final perturbation (i.e., increase the probability of misclassifying the perturbed input).
The latter is a problem-specific loss function, named \textit{Signature-Loss}, which is conceived to return a positive cost when the activation of a neuron is outside the signature range [$v_{\hat{y}, k, j}^{\text{min}}, v_{\hat{y}, k, j}^{\text{max}}$], where $\hat{y}$ is the label associated by the network to the input to be perturbed.
Technically speaking, the Signature-Loss is implemented as $\mathcal{L}_S = \sum_{n_{k,j} \in N}\text{ReLU}(v_{k,j} - v_{\hat{y}, k, j}^{\text{max}}) + \text{ReLU}(-v_{k,j} + v_{\hat{y}, k, j}^{\text{min}})$, where $\text{ReLU}(x) = max(0,x)$.}
{\NEWCHANGE
Figure~\ref{fig:loss_signature} provides a sample illustration of function $\mathcal{L}_S$ in the case of a single neuron only. 
The Signature-Loss has to be minimized during the optimization process to reduce the chance of the adversarial example being detected by the proposed CAMs.  

\begin{figure}
\centering
\makebox[\columnwidth]{\includegraphics[scale=0.34]{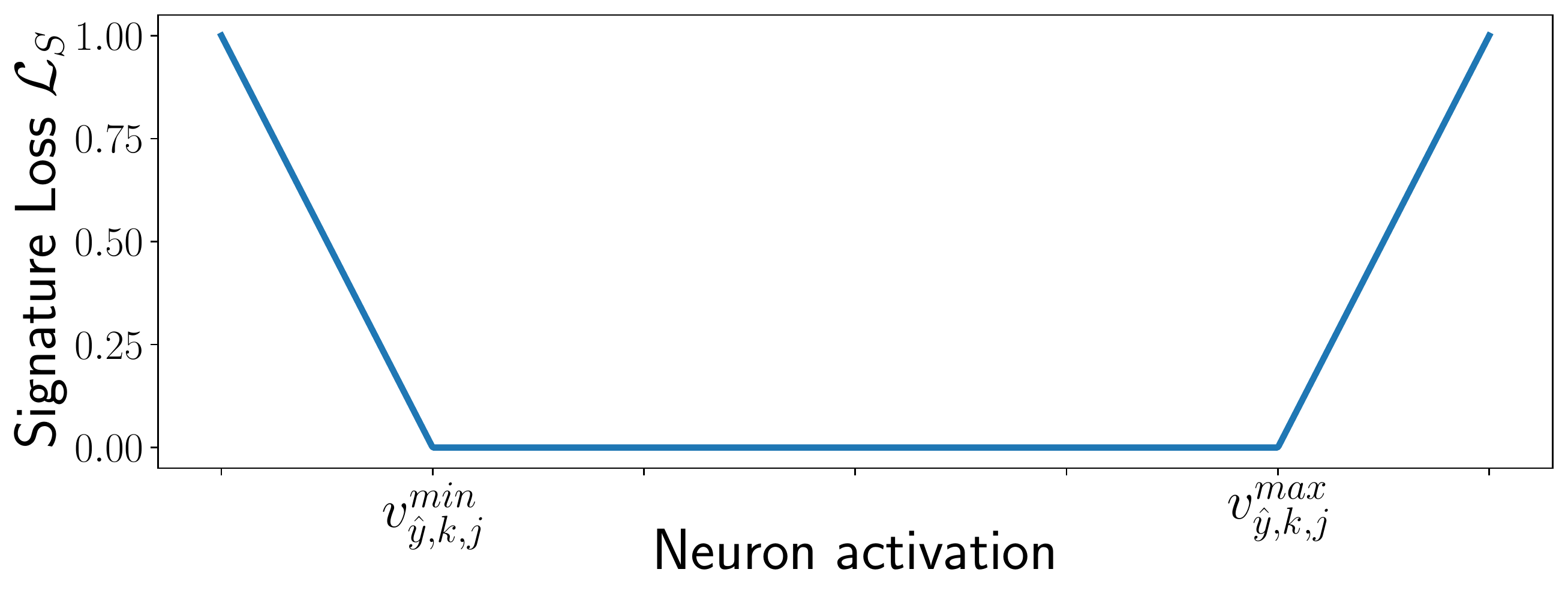}}
\vspace{-2.0em}
\caption{\NEWCHANGE Illustration of the \textit{Loss-Signature} for the case of a single neuron.}
\vspace{-1.2em}
\label{fig:loss_signature}
\end{figure}

From a practical point of view, the optimization problem to accomplish the Signature-Attack is solved through an iterative formulation similar to \cite{pgd_attack}, but based on a specific loss function $\mathcal{L}$ that takes into account both $\mathcal{L}_{CE}$ and $\mathcal{L}_{S}$:
\begin{equation}
  \begin{aligned}
    x_{k}^\prime = x_{k-1}^\prime + \alpha \cdot sign(\bigtriangledown_{x_{k-1}^\prime } \mathcal{L}(f(x_{k-1}^\prime ),\hat{y})), \\
    s.t. \quad ||x_{k}^\prime - x|| \leq \epsilon,
  \end{aligned}
\end{equation}
\noindent where $\alpha$ and $\epsilon$ denote the step size and the overall perturbation magnitude, respectively, as in \cite{pgd_attack}. 
The gradient of the loss function $\mathcal{L}$ is defined as
\begin{equation}
  \begin{aligned}
   \bigtriangledown_{x}\mathcal{L} = 
  (1 - \gamma) \cdot \frac{\bigtriangledown_{x}\mathcal{L}_{CE}}{||\bigtriangledown_{x}\mathcal{L}_{CE}||_2} - 
  \gamma \cdot
  \frac{\bigtriangledown_{x}\mathcal{L}_S}{||\bigtriangledown_{x}\mathcal{L}_S||_2},
  \end{aligned}
\end{equation}
where $\gamma$ is a parameter introduced to balance the importance of $\mathcal{L}_{CE}$ and $\mathcal{L_S}$. 
Since the gradients of $\mathcal{L}_{CE}$ and $\mathcal{L}_{S}$ may have very different scales, they are both subject to \textit{norm-2} normalization to ensure that their effect can be properly balanced using a single parameter $\gamma$.
Furthermore, it is also important to observe how the overall gradient of $\mathcal{L}$ goes in the same direction of $\bigtriangledown_{x}\mathcal{L}_{CE}$ and in the opposite direction of $\bigtriangledown_{x}\mathcal{L_{S}}$. Thus, optimizing the perturbation in the gradient direction of $\mathcal{L}$ means increasing the classification loss function $\mathcal{L}_{CE}$ while reducing the Signature-Loss $\mathcal{L_S}$.

\textbf{Effects of the Signature-Attack.}
Multiple versions of the Signature-Attack were tested on both MNIST and FMNIST. Similar to the settings used in Table~\ref{table:set_test}, we set $\alpha$ and $\epsilon$ to $0.004$ and $0.1$, respectively, and we evaluated five values for $\gamma$: $0.0$, $0.25$, $0.50$, $0.75$, and $1.0$.
The first and last values were tested for studying the behavior of the proposed attack in two limit cases (i.e., when only one of the two loss functions is considered).

Figure $\ref{fig:loss_test}$ reports the average classification loss $\mathcal{L}_{CE}$ and Signature-Loss $\mathcal{L}_S$ computed on the MNIST data set for the tested values of $\gamma$. 
Note that the classification loss (to be maximized) preserves a large value for the tested values of $\gamma < 1 $, while the Signature-Loss (to be minimized) has a considerably low value for the tested values of $\gamma \geq 0.5$.

The effectiveness of the generated adversarial examples can be observed from the results reported in Table \ref{tab:tableattacksignature} as a function of $\gamma$ (the results for $\gamma = 1.0$ are omitted since it was not possible to generate adversarial examples with this setting). 
As it can be noted from the table, the detection performance of the proposed CAMs is reduced, with respect to the results of Table~\ref{table:test}, on both MNIST and FMNIST, demonstrating the effectiveness of the Signature-Attack with $\gamma>0$.

\begin{figure}
\centering
\makebox[\columnwidth]{\includegraphics[scale=0.32]{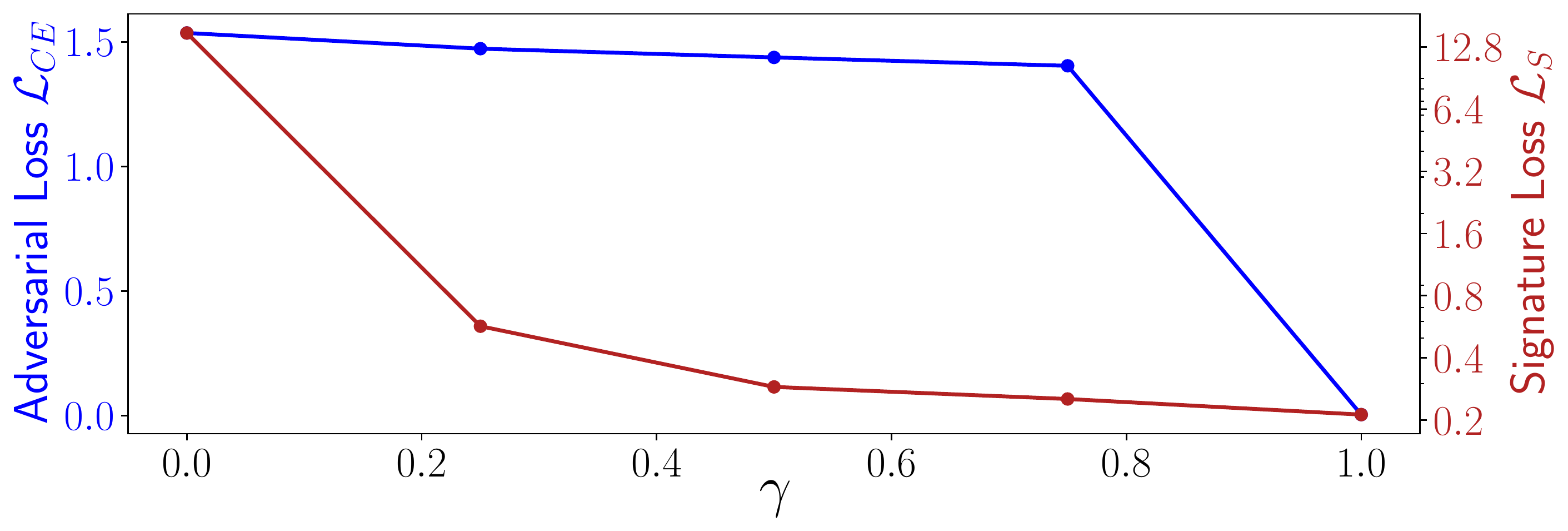}}
\vspace{-2.0em}
\caption{\NEWCHANGE Average losses computed with inputs from the MNIST data set when attacked with the Signature-Attack, under different configurations of the control parameter $\gamma$.}
\label{fig:loss_test}
\end{figure}

\begin{table}
\color{black}
\centering
\begin{tabular}{|c|ccc|ccc|}
\hline
         & \multicolumn{3}{c|}{\textbf{MNIST}}                               & \multicolumn{3}{c|}{\textbf{F-MNIST}}                             \\ \hline
$\gamma$ & \multicolumn{1}{c|}{SRC}   & \multicolumn{1}{c|}{MRC-32} & MRC-16 & \multicolumn{1}{c|}{SRC}   & \multicolumn{1}{c|}{MRC-32} & MRC-16 \\ \hline
0.0      & \multicolumn{1}{c|}{0.984} & \multicolumn{1}{c|}{0.957}  & 0.935  & \multicolumn{1}{c|}{0.968} & \multicolumn{1}{c|}{0.908}  & 0.912  \\ \hline
0.25     & \multicolumn{1}{c|}{0.779} & \multicolumn{1}{c|}{0.775}  & 0.701  & \multicolumn{1}{c|}{0.881} & \multicolumn{1}{c|}{0.805}  & 0.785  \\ \hline
0.50     & \multicolumn{1}{c|}{0.756} & \multicolumn{1}{c|}{0.769}  & 0.690  & \multicolumn{1}{c|}{0.879} & \multicolumn{1}{c|}{0.797}  & 0.781  \\ \hline
0.75     & \multicolumn{1}{c|}{0.749} & \multicolumn{1}{c|}{0.742}  & 0.687  & \multicolumn{1}{c|}{0.874} & \multicolumn{1}{c|}{0.793}  & 0.778  \\ \hline
\end{tabular}

\caption{Detection accuracy on FMNIST and MNIST against adversarial examples crafted with the Signature-Attack under four configurations.}
\label{tab:tableattacksignature}
\vspace{-1em}
\end{table}

\textbf{Countermeasure.}
To counteract the Signature-Attack, we included a portion of the adversarial examples crafted with that attack to the calibration set in order to re-calibrate the thresholds.
This approach helps improve the mechanism by tuning the thresholds at lower values such that also CAM-specific adversarial inputs can be detected without significantly reducing the accuracy of all the others. 

To this purpose, we refined the set of inputs used for calibration (see Section~\ref{s:exp_calib}) by adding 800 adversarial examples obtained by the Signature-Attack with $\gamma \in \{ 0.25,0.75\}$.
Other $8000$ adversarial examples ($2000$ for each $\gamma \in \{ 0.0, 0.25, 0.50, 0.75\}$) are appended to the test set introduced in Table~\ref{table:set_test}. 
Table~\ref{table:final_test} shows the final results obtained after the re-calibration, i.e., by testing the detection mechanisms with the new thresholds.
By comparing Table~\ref{table:final_test} with Table~\ref{tab:tableattacksignature}, it can be noted that the detection accuracy of the proposed CAMs is improved for all the tested versions of the Signature-Attack. Also note that the re-calibration of thresholds did not significantly affect the detection performance for the other kinds of unsafe inputs, which shows similar results with respect to Table~\ref{table:test}.

\begin{table*}
\color{black}
\centering
\resizebox{0.9\linewidth}{!}{%
\begin{tabular}{|c|cccc|cccc|}
\hline
 & \multicolumn{4}{c|}{MNIST} & \multicolumn{4}{c|}{F-MNIST} \\ \hline
Input type & \multicolumn{1}{c|}{SRC} & \multicolumn{1}{c|}{MRC-32} & \multicolumn{1}{c|}{MRC-16} & kNNC & \multicolumn{1}{c|}{SRC} & \multicolumn{1}{c|}{MRC-32} & \multicolumn{1}{c|}{MRC-16} & kNNC \\ \hline
FGSM-1 & \multicolumn{1}{c|}{1.0} & \multicolumn{1}{c|}{0.986} & \multicolumn{1}{c|}{0.962} & 0.969 & \multicolumn{1}{c|}{0.839} & \multicolumn{1}{c|}{0.845} & \multicolumn{1}{c|}{0.829} & 0.922 \\ \hline
FGSM-2 & \multicolumn{1}{c|}{1.0} & \multicolumn{1}{c|}{0.995} & \multicolumn{1}{c|}{0.991} & 0.951 & \multicolumn{1}{c|}{0.925} & \multicolumn{1}{c|}{0.948} & \multicolumn{1}{c|}{0.922} & 0.891 \\ \hline
PGD-1 & \multicolumn{1}{c|}{0.991} & \multicolumn{1}{c|}{0.953} & \multicolumn{1}{c|}{0.887} & 0.969 & \multicolumn{1}{c|}{0.849} & \multicolumn{1}{c|}{0.842} & \multicolumn{1}{c|}{0.804} & 0.939 \\ \hline
PGD-2 & \multicolumn{1}{c|}{1.0} & \multicolumn{1}{c|}{0.977} & \multicolumn{1}{c|}{0.960} & 0.953 & \multicolumn{1}{c|}{0.961} & \multicolumn{1}{c|}{0.889} & \multicolumn{1}{c|}{0.874} & 0.868 \\ \hline
BIM-1 & \multicolumn{1}{c|}{0.993} & \multicolumn{1}{c|}{0.910} & \multicolumn{1}{c|}{0.953} & 0.923 & \multicolumn{1}{c|}{0.839} & \multicolumn{1}{c|}{0.847} & \multicolumn{1}{c|}{0.792} & 0.919 \\ \hline
BIM-2 & \multicolumn{1}{c|}{0.999} & \multicolumn{1}{c|}{0.967} & \multicolumn{1}{c|}{0.940} & 0.949 & \multicolumn{1}{c|}{0.980} & \multicolumn{1}{c|}{0.921} & \multicolumn{1}{c|}{0.924} & 0.844 \\ \hline
CW & \multicolumn{1}{c|}{0.987} & \multicolumn{1}{c|}{0.977} & \multicolumn{1}{c|}{0.953} & 0.960 & \multicolumn{1}{c|}{0.813} & \multicolumn{1}{c|}{0.825} & \multicolumn{1}{c|}{0.822} & 0.907 \\ \hline
Out of Dis. & \multicolumn{1}{c|}{1.0} & \multicolumn{1}{c|}{1.0} & \multicolumn{1}{c|}{1.0} & 0.34 & \multicolumn{1}{c|}{1.0} & \multicolumn{1}{c|}{1.0} & \multicolumn{1}{c|}{1.0} & 0.851 \\ \hline
Patch & \multicolumn{1}{c|}{1.0} & \multicolumn{1}{c|}{0.889} & \multicolumn{1}{c|}{0.921} & 0.988 & \multicolumn{1}{c|}{1.0} & \multicolumn{1}{c|}{0.907} & \multicolumn{1}{c|}{0.853} & 0.954 \\ \hline
\multicolumn{1}{|l|}{Signature-Attack ($\gamma=0.0$)} & \multicolumn{1}{c|}{0.998} & \multicolumn{1}{c|}{0.896} & \multicolumn{1}{c|}{0.841} & 0.973 & \multicolumn{1}{c|}{0.963} & \multicolumn{1}{c|}{0.889} & \multicolumn{1}{c|}{0.884} & 0.851 \\ \hline
\multicolumn{1}{|l|}{Signature-Attack ($\gamma=0.25$)} & \multicolumn{1}{c|}{0.949} & \multicolumn{1}{c|}{0.939} & \multicolumn{1}{c|}{0.782} & 0.972 & \multicolumn{1}{c|}{0.935} & \multicolumn{1}{c|}{0.861} & \multicolumn{1}{c|}{0.852} & 0.852 \\ \hline
\multicolumn{1}{|l|}{Signature-Attack ($\gamma=0.50$)} & \multicolumn{1}{c|}{0.939} & \multicolumn{1}{c|}{0.831} & \multicolumn{1}{c|}{0.767} & 0.971 & \multicolumn{1}{c|}{0.933} & \multicolumn{1}{c|}{0.861} & \multicolumn{1}{c|}{0.852} & 0.853 \\ \hline
\multicolumn{1}{|l|}{Signature-Attack ($\gamma=0.75$)} & \multicolumn{1}{c|}{0.925} & \multicolumn{1}{c|}{0.820} & \multicolumn{1}{c|}{0.763} & 0.971 & \multicolumn{1}{c|}{0.931} & \multicolumn{1}{c|}{0.858} & \multicolumn{1}{c|}{0.847} & 0.851 \\ \hline
Safe Samples & \multicolumn{1}{c|}{0.986} & \multicolumn{1}{c|}{0.938} & \multicolumn{1}{c|}{0.89} & 0.957 & \multicolumn{1}{c|}{0.890} & \multicolumn{1}{c|}{0.886} & \multicolumn{1}{c|}{0.854} & 0.916 \\ \hline
\end{tabular}%
}
\caption{Detection accuracy after the re-calibration, performed with also unsafe inputs crafted from the proposed Signature-Attack.}
\label{table:final_test}
\end{table*}

}

\section{Conclusion and future work} \label{s:conc}

This paper presented a monitoring architecture to enhance the trustworthiness of DNNs by generating a confidence value to be coupled to the network prediction. The workflow of the proposed architecture includes an off-line phase (to be performed after training) where a DNN Signature is generated starting from a set of trusted inputs,and an online phase (to be performed at inference time), where the DNN Active State generated by a new input is compared against the DNN Signature.
Four different CAMs have been proposed to instantiate the proposed architecture.
An implementation as an extension of the Caffe framework has also been presented.

{\CHANGE
The proposed CAMs have been compared in terms of detection performance and running time, using several state-of-the-art methods to generate unsafe inputs. The experimental results on MNIST and F-MNIST data sets showed that the CAMs based on coverage techniques better generalize among different types of unsafe inputs. VisionGuard \cite{vision_guard} and FeatureSqueezing \cite{feature_squeezing}, which proved to have competitive performance with respect to most of the proposed CAMs for low-perturbation AEs, lose their effectiveness with medium-perturbation attacks and out-of-distribution unsafe inputs. 
Also in terms of running time and memory footprint, 
most of the proposed CAMs exhibit better performance than previous work, proving to be more suitable for real-time applications.  

The approach proposed in this paper opens several interesting future work directions. Among them, a particularly relevant research line is the investigation of techniques that combine the proposed CAMs to enhance the detection performance 
while balancing the implied additional inference time and memory footprint.
Research efforts should also be spent on the design of other CAMs and techniques to prune unnecessary information stored by the DNN Signature.
Such developments will hopefully make possible to apply the proposed approach to more complex DNNs, such as ResNet~\cite{resnet} and Inception~\cite{inception}, and larger and sparser data sets, as ImageNet~\cite{imagenet}.
}

\bibliographystyle{IEEEtran}
\bibliography{IEEEabrv, main}

\clearpage
\appendices
\pagenumbering{gobble}
\twocolumn[
  \begin{@twocolumnfalse}
    \begin{center}
      \Huge
      \textsf{Supplementary material for the paper ``Increasing the Confidence of Deep Neural Networks
by Coverage Analysis''}
    \end{center}
    \vspace{1pt}
    \begin{center}
      \large
      \textup{
      \textsf{Giulio Rossolini, Alessandro Biondi, Giorgio
    Buttazzo}}
    \end{center}
    \vspace{10pt}
  \end{@twocolumnfalse}
]
\section{Neuron Rank Coverage (NRC)}
This section presents a fourth CAM based on the Top-level Neuron coverage criterion~\cite{DeepGauge}, which is quite different from the previous ones.
It is based on determining a ranking of the neurons based on their output values and then recording how often a neuron appears in the top positions of the ranking.

Given a number of positions $P \geq 1$, during the offline phase, the Trusted Set is processed to build a ranking of the neuron outputs for each class, and then counting how many times a neuron ends up in one of the top $P$ positions of its corresponding ranking over all inputs in the Trusted Set. This information is then used to build the DNN Signature.

Formally speaking, given a network input $x$, a ranking $R(x)_k^h$ is introduced to denote the neurons of the $h$-th channel of the $k$-th layer, i.e., $N_k^{\text{ch-}h}$, sorted by descending neuron outputs $v_{k,j}(x)$.
Then, a function $\textit{rank}_{k,j}(x)$ is defined to denote the position (numbered starting from 1) of neuron $n_{k,j} \in N_k^{\text{ch-}h}$ within the ranking $R(x)_k^h$.

The DNN Signature $\sigma_i$ for the $i$-th class is a collection of frequencies $\sigma_{i,k,j} = \lambda_{i,k,j}^{\text{top-P}} \in [0, 1]$, one for each neuron, given by the number of times the neuron ends up in one of the top $P$ positions of its ranking when the network input is in $S_i$, divided by the cardinality of $S_i$ itself, i.e.,
\begin{equation}
\lambda_{i,k,j}^{\text{top-P}} = \frac{ |x \in S_{i} \: :  \: \textit{rank}_{k,j}(x) \leq P |}{|S_{i}|}. 
\end{equation}

These frequencies are computed by the Aggregation Algorithm of NRC, which is reported in Algorithm~\ref{alg:agg-NRC}.

\begin{algorithm}[htb!]
\SetAlgoLined
 \textbf{Input} Trusted Set $S$, trained DNN, reference ranking $P$ \\
 \textbf{Output} DNN Signature $\sigma$ \\
 \algrule
 \For{$S_{i} \in S$} { 
 $\sigma_i = \{ \}$\\
 $\forall n_{k,j} \in N, ~~ \lambda_{i,k,j}^{\text{top-P}}=0$\\
 	\For{$x \in S_i$}{
 		\For{$L_k \in L$}{
 		\For{$h=1, \ldots, C_k$}{
 		Compute ranking $R(x)_k^h$\\
 		\For{$n_{k,j} \in N_k^h$} {
			 \If{$\text{rank}_{k,j}(x) \leq P$}{
			 $\lambda_{i,k,j}^{\text{top-P}}$++
			 }		
 		}
 		}
 		}
 	}
 	$\forall n_{k,j} \in N, ~~ \lambda_{i,k,j}^{\text{top-P}} = \frac{\lambda_{i,k,j}^{\text{top-P}}}{|S_i|}$\\
 	$\forall n_{k,j} \in N$, add $\sigma_{i,k,j} = \lambda_{i,k,j}^{\text{top-P}}$ to $\sigma_i$
 }
 \Return $\sigma = \{\sigma_1, \ldots, \sigma_{m} \}$ \\
\caption{Aggregation Algorithm of NRC.}
\label{alg:agg-NRC}
\end{algorithm}

During the online phase of NRC, given a new input $x_{\text{new}}$, all rankings $R(x_{\text{new}})_k^h$ are computed 
to in turn compute function $\textit{rank}_{k,j}(x_{\text{new}})$ for all neurons. This allows obtaining the DNN Active State.
A cost $\Theta_{k,j}(x_{\text{new}}) \in [0, 1]$ is defined for each neuron $n_{k,j}$ to quantify how much the neuron output is deemed safe as a function of the DNN Active State. Being $\hat{y}$ the index of the class assigned to $x_{\text{new}}$, this cost is formally defined as
\begin{equation}
\Theta_{k,j}(x_{\text{new}}) = 
\begin{cases} 
\sigma_{\hat{y},k,j}	, & \mbox{if }\textit{rank}_{k,j}(x_{\text{new}}) \leq P \\ 
0, & \mbox{otherwise.}
\end{cases}
\end{equation}

The interpretation of the cost is the following: if a neuron is in the top-$P$ positions of its ranking, then it is deemed relevant for the network output and its contribution to the quantification of the overall safety of $x_{\text{new}}$ is given by the frequency with which it has been ranked among the top-$P$ neurons when testing the Trusted Set.
In this way, neurons with a high ranking in the DNN Active State that are also frequently ranked among the top-$P$ ones in the DNN Signature, are treated as indicators of a tested activation pattern of the network, hence positively contributing to deeming $x_{\text{new}}$ safe.
These costs are computed by the Confidence Evaluation Algorithm and the overall sum is denoted as $\eta$. For this CAM, the confidence score is computed as $c = 1 - \exp({- \frac{\eta \cdot ln(2)}{\tau_{\hat{y}}}})$.
Such a procedure is reported in Algorithm~\ref{alg:safety-eval-NRC}.

\begin{algorithm}[htb!]
\SetAlgoLined
 \textbf{Input} input $x_{new}$, DNN Signature $\sigma$, trained DNN, thresholds $\tau_i$, reference ranking $P$ \\
 \textbf{Output} confidence $c$ \\
 \algrule
 $\hat{y} = \text{argmax}_{1\leq j \leq m} \{ f(x_{\text{new}})\}$\\
 $\eta$ = $0$ \\
 \For{$L_k \in L$}{
 \For{$h=1, \ldots, C_k$}{ 
 Compute ranking $R(x_{\text{new}})_k^h$\\
 \For{$n_{k,j} \in N_k^h$} {
 Extract $\sigma_{\hat{y}, k, j}$ from $\sigma_{\hat{y}}$\\
 \If{$\textit{rank}_{k,j}(x_{\text{new}}) \leq P $}  {
 $\eta$ += $\sigma_{\hat{y}, k, j}$
 }
 
 }
 }
 }
 \Return $c = 1 - \exp({- \frac{\eta \cdot ln(2)}{\tau_{\hat{y}}}})$ \\
 
 \caption{Confidence Evaluation Algorithm of NRC.}
 \label{alg:safety-eval-NRC}
\end{algorithm}

\textbf{Experimental Evaluation of NRC.}
While the detection performance of NRC is not satisfactory, we report that the percentage increase of the inference time introduced by NRC ranges from 1102$\%$ to 40.2$\%$, corresponding to the installations denoted by $\{$1,2,3$\}$ and $\{$3$\}$, respectively. Nevertheless, it must be noted that the extraction of the Active State required by NRC has been executed on CPU as it requires a sorting operation (to build the neuron ranks) that is hard to efficiently implement on GPU.

\section{Coverage Testing for DNNs}
This section aims at clarifying the main differences between coverage methods for testing DNNs and the proposed CAMs for run-time monitoring. Subsequently, we provide a brief description of the original coverage techniques that inspired the design of our CAMs.

\subsection{Coverage-based Testing vs. Runtime Monitoring}
As for classical software, coverage testing for DNNs~\cite{survey} consists of a set of techniques aimed at evaluating the trustworthiness of a given model with respect to testing inputs. 
As such, coverage testing is intended to be performed during the testing phase of the DNN, where the model robustness is on trial to accept or reject its deployment.
The main questions that DNN coverage testing aims at answering are: Is the test set enough for evaluating the model? How much the test set is capable of stimulating all possible network behaviors? 
Coverage testing is accomplished by defining a set of coverage metrics that establish a way to identify internal model patterns that can be covered or not by a given test set.  Finally, a coverage evaluation is returned to quantify how much the test set stimulates all the internal DNNs activation patterns.
Various test-generation techniques, such as~\cite{DeepTest},~\cite{mdcddnn1}, have been proposed for increasing the coverage of a DNN. 
They leverage patterns that were left uncovered by the original test set (i.e., the one used to assess the performance of the DNNs after training) for crafting new inputs that cover them.

\begin{figure}[!t]
\centering
\makebox[\columnwidth]{\includegraphics[scale=0.58]{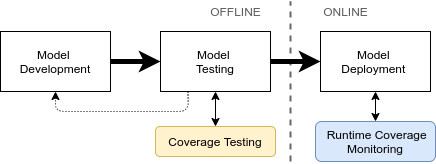}}
\vspace{-1.5em}
\caption{\small{Differences between coverage testing and run-time coverage monitoring in the production pipeline of DNNs.}}
\vspace{-1.2em}
\label{fig:cov_and_cov}
\end{figure}

This work leverage coverage metrics for recording safe activation patterns offline to be then compared online against those generated by new inputs provided to the network. As such, this work operates in the deployment phase of the DNN production pipeline, while previous work on coverage addressed the testing phase.
Figure \ref{fig:cov_and_cov} summarizes the previous observations.  

\subsection{Definition of original Coverage metrics}
Nevertheless, the proposed CAMs take inspiration from original metrics used in coverage testing. In particular, the considered metrics were designed for identifying DNN features (e.g., neuron activation values) that are deemed covered if it exists at least one inputs $x$ in a given test set $\mathcal{T}$ that meets certain conditions specified by the coverage metric of interest. 
As such, these metrics, in their original formulation, are not compliant with the run-time monitoring scheme presented in Section 4 of the paper.
For the sake of completeness, their definitions as provided by \cite{DeepGauge} and ~\cite{survey} are reported next, using the notation introduced in Section 3 of the paper. 

\textbf{Neuron coverage.} \textit{A neuron $n_{k,i}$ is covered by a test input $x$
if $v_{k,i}$ is larger than a certain threshold.}
Such a threshold can be manually selected or, as usual, set equal to the maximum activation value obtained among all the inputs of a given set (e.g., the training set). 

\textbf{Multi-Section Coverage.} \textit{Given $m \geq 1$ sections within the domain of the neuron output, a neuron $n_{k,i}$ is covered by a test set $\mathcal{T}$ if all m neurons sections are activated by inputs in $\mathcal{T}$.}
The original definition of the metric defined the sections by evenly splitting the range $[v^l_{k,i}, v^u_{k,i}]$, where $v^l_{k,i}$ and $v^u_{k,i}$ are the lower and upper bounds defined in Section 5 of the paper. 

\textbf{Top-level Neuron Coverage.} \textit{A neuron $n_{k,j}$ is covered by an input $x$ if $rank_{k,j} \leq m$. In other words the input $x$ covers $n_{k,j}$ if brings it to the top $m$ neurons activated by $x$.}

\section{Experimental evaluation}

This section provides additional information on the experimental evaluation discussed in 
Section 7 of the paper.

\subsection{Generation of unsafe inputs}
\begin{figure*}[!t]
\centering
  \subfloat[\label{1a}]{%
       \includegraphics[width=0.14\linewidth]{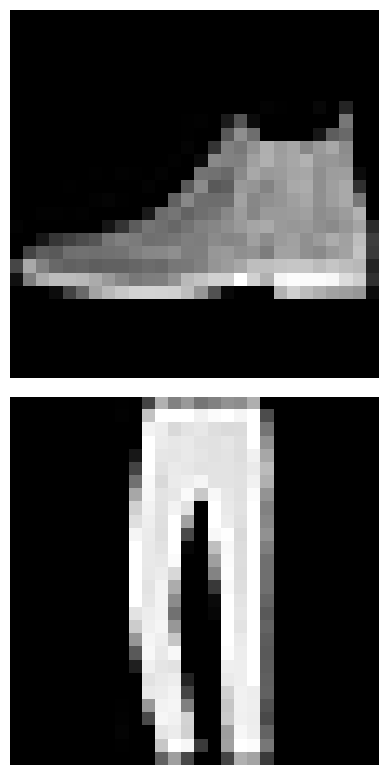}}
    \hfill
  \subfloat[\label{1b}]{%
        \includegraphics[width=0.14\linewidth]{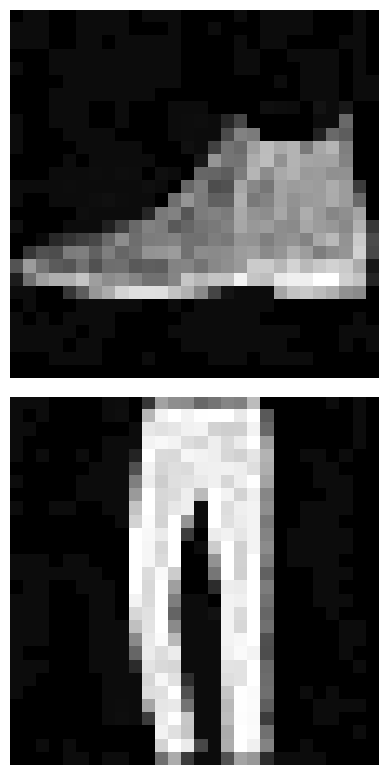}}
     \hfill
  \subfloat[\label{1b}]{%
        \includegraphics[width=0.14\linewidth]{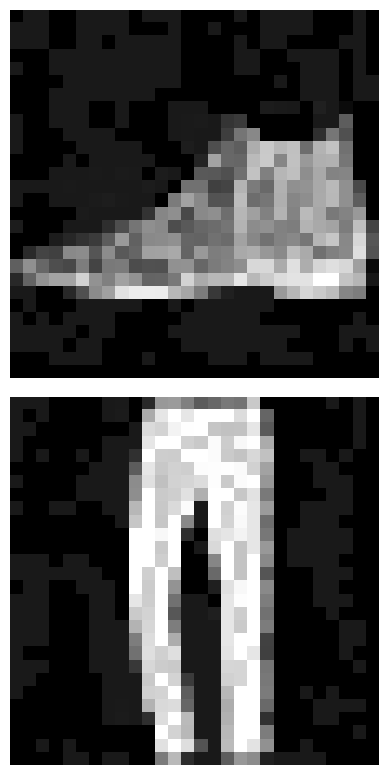}}
     \hfill
  \subfloat[\label{1b}]{%
        \includegraphics[width=0.14\linewidth]{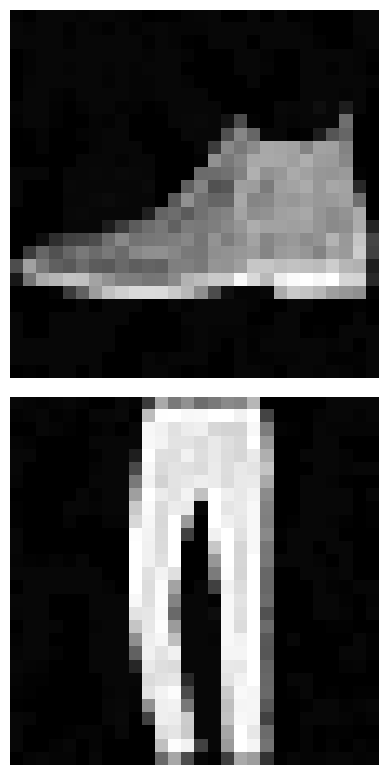}}
     \hfill
  \subfloat[\label{1b}]{%
        \includegraphics[width=0.14\linewidth]{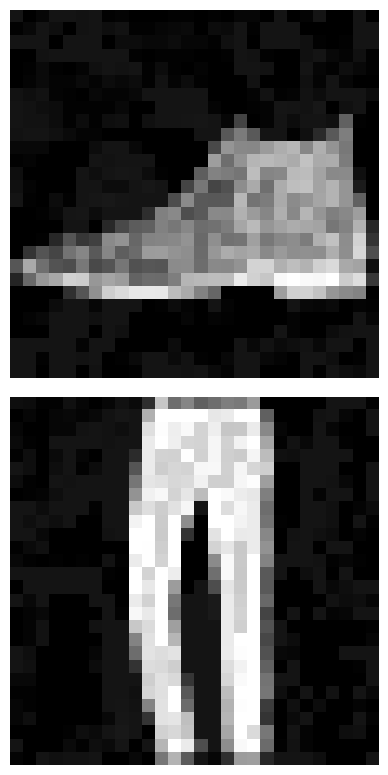}}
     \hfill
  \subfloat[\label{1b}]{%
        \includegraphics[width=0.14\linewidth]{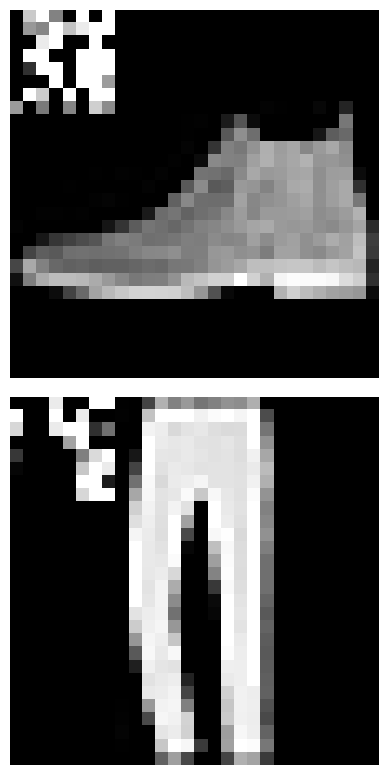}}
     \hfill
  \subfloat[\label{1b}]{%
        \includegraphics[width=0.14\linewidth]{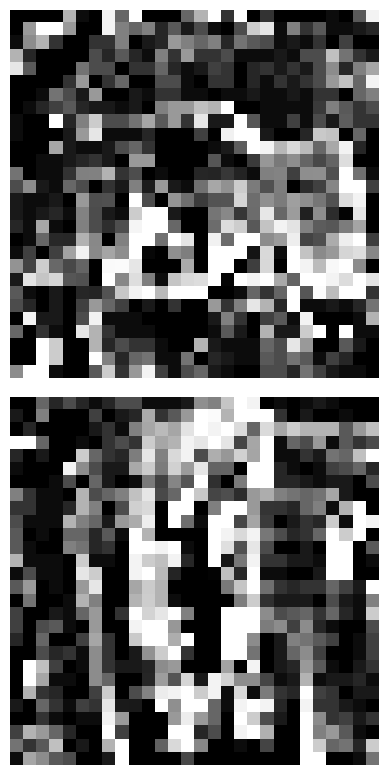}}
        \vspace{-0.5em}
        \caption{Examples of a perturbed image with adversarial attacks. (a) is the clear image; (b)(c) are the obtained AEs using a FGSM with $\epsilon$ equal to 0.05 and 0.1, respectively. (d)(e) are crafted using a PGD with $\epsilon$ equal to 0.3 and 0.1. Finally (f) and (g) represent unsafe inputs obtained from the illustrated patch attack and out-of-distribution attack.}
        \vspace{-0.5em}
        \label{fig:aes}
\end{figure*}

The following paragraphs provide a detailed description of the generation methods used for crafting all the tested unsafe inputs. Figure~\ref{fig:aes} illustrates some of those executed on images from the FMNIST dataset. 
Note that all the crafted unsafe inputs are kept in the original image domain, i.e. all the image pixels are $\in [0,1]$ after normalization.

\textbf{FGSM.} The Fast Gradient Sign Method (FGSM)~\cite{goodfellow2014explaining} is one of the first white-box methods proposed for generating AEs. Given an input $x$ and its label encoded as a one-hot vector $y$, the input is manipulated by adding a small perturbation $\epsilon$ 
in the direction of the gradient of the loss function $\bigtriangledown_x{} \mathcal{L}( f(x),y)$, which is obtained by its sign.
Formally, adversarial inputs $x^\prime$ are generated as:
\begin{equation}
x^\prime=x+\epsilon\cdot{}sign(\bigtriangledown_x{} \mathcal{L}( f(x),y)).
\end{equation}
Modifying an input in the direction of the gradient has the effect of increasing the classification error. 
This method is straightforward and efficient, and it is able to create a large set of AEs with different perturbation levels by slightly changing the $ \epsilon $ parameter.

\textbf{PGD.}
The FGSM is a one-step attack that aims at generating a single perturbation of the input.
Better results can be obtained by refining such a perturbation by solving an optimization problem.
Indeed, other attack techniques rely on a constrained optimization method for finding a perturbation that maximizes the loss function related to the correct label $y$, i.e., $max_{ x^\prime }(\mathcal{L}( f(x^\prime),y))$, and constraining the search in a $\epsilon$-neighborhood around $x$, such that $|| x -x^\prime||_{p} \leq \epsilon$, where $\epsilon$ is a problem parameter for tuning the perturbation and $p$ is a mathematical norm. 
The Projected Gradient Descent (PGD) is a popular iterative method that can be applied to solve constrained optimization problems using a step size $\alpha$. Madry et al.~\cite{pgd_attack} have extended the PGD for computing AEs, solving the above optimization problem.

\textbf{CW.} Carlini and Wagner~\cite{CarliniAttack} proposed an attack scheme based on distance metrics. The method relies on solving an optimization problem that minimizes the perturbation applied to the original samples and, at the same time, maximizes the probability of matching a target class label. Notably, this optimization scheme is one of the most efficient attacks (among the white-box ones) to generate AEs that are very close to the corresponding original inputs. The tested AEs are crafted using $c = 0.0001$ and $\kappa = 0$ that are both specific parameters of the CW optimization problem.

\textbf{BIM.}
It is an iterative version of the FGSM presented by Kurakin et al.~\cite{BIM}. 
The basic idea is to apply multiple times the adversarial perturbation with a smaller step $\alpha$, 
hence generating a series of refined AEs $x_{0}^\prime, x_{1}^\prime, x_{2}^\prime, \ldots$, with the following 
formula:
\begin{equation}
x_{0}^\prime =x, \quad
x_{k}^\prime = clip_{x, \epsilon} (x_{k-1}^\prime + \alpha \cdot{}sign(\bigtriangledown_{x_{k-1}^\prime} \mathcal{L}( f(x_{k-1}^\prime ),y))),
\end{equation}

\noindent where $clip_{x,\epsilon}$ represents a saturation operator that bounds its argument within an $\epsilon$-neighborhood  (i.e., $\forall a,  clip_{x,\epsilon}(a) \leq ||x + \epsilon||_{p} $). Compared to FGSM, it provides a finer  control of the generation process that allows creating AEs with very small perturbations.

\textbf{Out-of-distribution unsafe inputs.}
Although the trustworthiness of DNNs was mainly studied with respect to AEs in previous work, in practice 
there are other types of inputs that could cause a DNN to produce a wrong prediction.
For instance, they may be inputs that have no particular meaning from a human perspective but lead the network to 
a certain prediction with a high confidence value due to over-generalization.
As such, they must detected.

We have generated a large set of such unsafe inputs in order to evaluate the capability of the algorithms to detect them at runtime.
This kind of inputs were obtained using a \textit{multi-step targeted FGSM} method starting from 
safe inputs.
Given a target label $y_{\text{target}}$ to be assigned by the network to one of such unsafe inputs, this approach can be formulated as: 
\begin{equation}
x_{k}^\prime = x_{k-1}^\prime - \frac{\epsilon}{(k+1)^{2}} \cdot sign(\bigtriangledown_{x_{k-1}^\prime } \mathcal{L}(f(x_{k-1}^\prime ),y_{\text{target}})),
\end{equation}
\noindent where $x_{0}^\prime = x$.
The idea is to move the perturbation in the opposite direction of the loss function gradient related to the input $x$ and the target label $y_{\text{target}}$. 
In this way, the perturbation leads the network to increase the confidence score of the target class. 
Note that, differently by the others iterative adversarial attacks presented above, here the perturbation is not clipped at each iteration within $\epsilon$. In fact, the step size of the perturbation is scaled proportionally with $k$, such that the method generates large perturbations during the first step (possibly reducing the human meaning of the image), and then smaller perturbations in the last ones to consolidate the prediction of the model with the target class. 
In this work, this iterative generation has been stopped when obtaining input samples that the network associates to $y_{\text{target}}$ with a corresponding softmax score higher than or equal to $0.99$.
Very often, the generated inputs tend to lose their human interpretation, as shown in Figure~\ref{fig:aes}.

\textbf{Adversarial Patches}
An adversarial patch consists of an image, with smaller size than the input $x$, realized to fool the model when added to $x$ in a certain position. Formally, let $\delta$ denote a patch and $\tilde{x}$ a patched image that is obtained by substituting pixels of $x$ with $\delta$ in an area of $x$ with the same size of $\delta$. 
The adversarial crafting process is similar to the previous ones, with the main difference that here the perturbation is applied at each iteration to the patched pixels only: 
\useshortskip
\begin{equation}
    \delta_{t} = \mathrm{clip}_{[0,1]} \left(\delta_{t-1} + \epsilon \bigtriangledown_{\delta_{k-1} }\mathcal{L}(f(\tilde{x}_{t-1}), y) \right ),
\end{equation} \label{e:opt}
\noindent where $\delta_{t}$ is the adversarial patch computed a time t and $\tilde{x_t}$ is the corresponding patched image obtained from $x$ and $\delta_{t}$.
The tested patches have a size of $8\times8$ pixels and are always placed at the top-left corner of the image. The size and position are decided to not overlapping the main content of images in MNIST and FMNIST (see Fig.~\ref{fig:aes}).

\smallskip
\textbf{Thresholds for dealing with unsafe inputs.}
As reported in Section 7.2 of the paper, a prediction is considered to be wrong when the DNN classifies such inputs in a wrong class with a prediction score larger than $0.8$, for AEs and adversarial patches, or larger than $0.99$, for out-of-distribution inputs.
Note that a lower bound for the softmax score produced by the network may help filter out several spurious predictions, even those that correspond to unsafe inputs. As we wanted to perform experiments with effective unsafe inputs, 
we recorded the mean and variance of the softmax scores produced by the network for the correctly-classified inputs in the test set, hence making sure to deal with unsafe inputs that are (incorrectly) classified with softmax scores comparable to the ones of safe inputs.
Nevertheless, note that the higher the softmax score with which AEs are (incorrectly) classified the more difficult their generation.
Hence, the selection of very high thresholds would imply very low likelihoods to generate a sufficient number of AEs to perform a meaningful experimental evaluation.
By empirically exploring this trade-off, we ended up in setting the threshold to $0.8$ for deeming wrong the prediction of the network when tested with AEs and adversarial patches.
Conversely, the situation for out-of-distribution unsafe inputs is quite different. Indeed, we found that it is much easier to generate such a kind of unsafe inputs that are (incorrectly) classified with very high softmax scores. For this reason, we set the threshold to $0.99$ when testing them.
Finally, it is worth observing that different thresholds can be used for our experimentation as we know the type of unsafe input with which we are testing the network. This choice allowed collecting more expressive experimental results as we can individually test strong attacks for each type of unsafe inputs.

\subsection{Threshold calibration and ROC Analysis} 
This experimental evaluation was focused on a binary classification of the network inputs, i.e., either an input was deemed \emph{safe} 
and the prediction made by the network was accepted, or the input was deemed \emph{unsafe}, and the network prediction was rejected.
This has been implemented by calibrating the tolerances $\tau_i$ used by the various Confidence Evaluation Algorithms so that 
an input is deemed safe if $c \geq 0.5$ and unsafe otherwise.

As typical for binary classification problems, each input can be of one of the following types:
\begin{itemize}
\item False Positive (FP): it is classified as unsafe but it is safe;
\item False Negative (FN): it is classified as safe but it is unsafe;
\item True Positive (TP): it is classified as unsafe and it is unsafe;
\item True Negative (TN): it is classified as safe and it is safe.
\end{itemize} 

The False Positive Rate (FPR) is defined as the ratio of safe inputs classified as unsafe, 
i.e., $\text{FPR} = \text{\#FP}/(\text{\#FP}+\text{\#TN})$, whereas the True Positive Rate (TPR) is defined as the ratio of unsafe inputs classified as unsafe, 
i.e., $\text{TPR} = \text{\#TP}/(\text{\#TP}+\text{\#FN})$.
An ideal binary classifier does not classify any safe input as unsafe (FPR$=0$) and classifies all unsafe inputs as unsafe (TPR$=1$).

\begin{figure*}
\centering
  \subfloat[\label{fig:roc_mnist}]{%
       \includegraphics[width=0.9\linewidth]{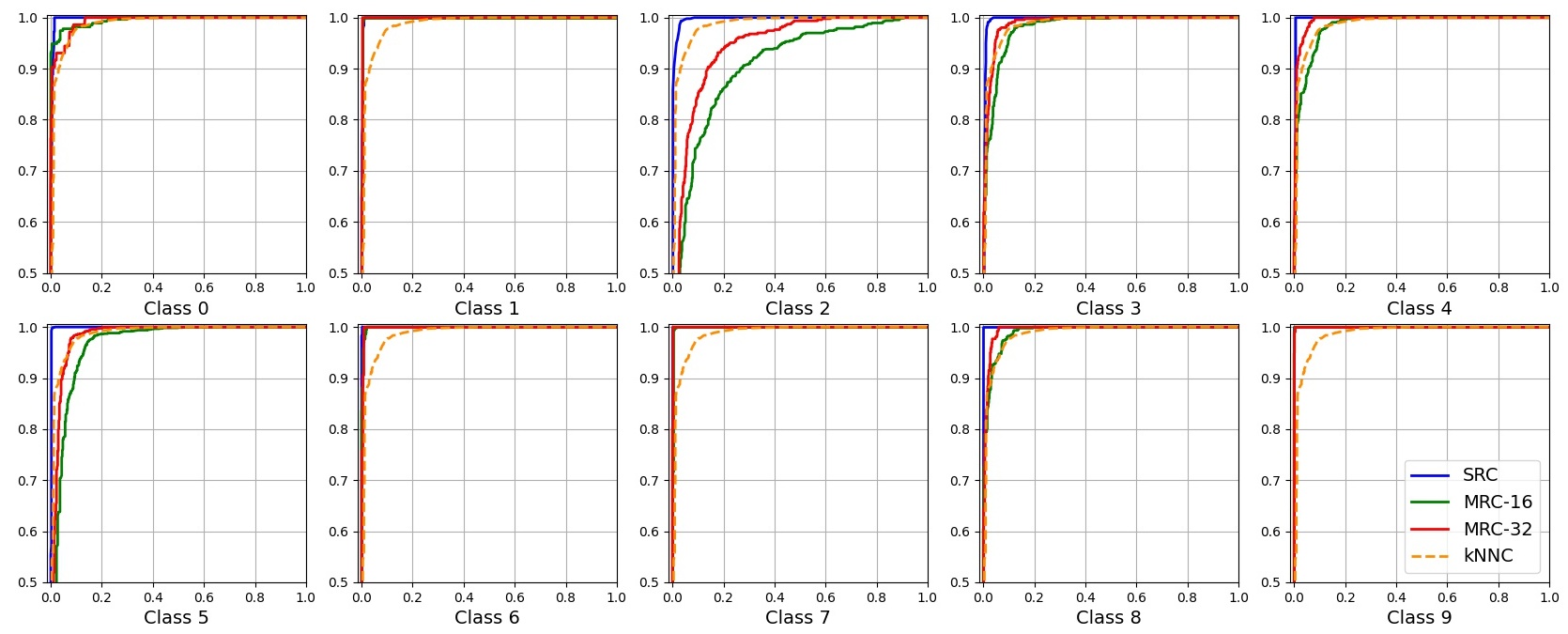}}
    \hfill
  \subfloat[\label{fig:roc_fmnist}]{%
        \includegraphics[width=0.9\linewidth]{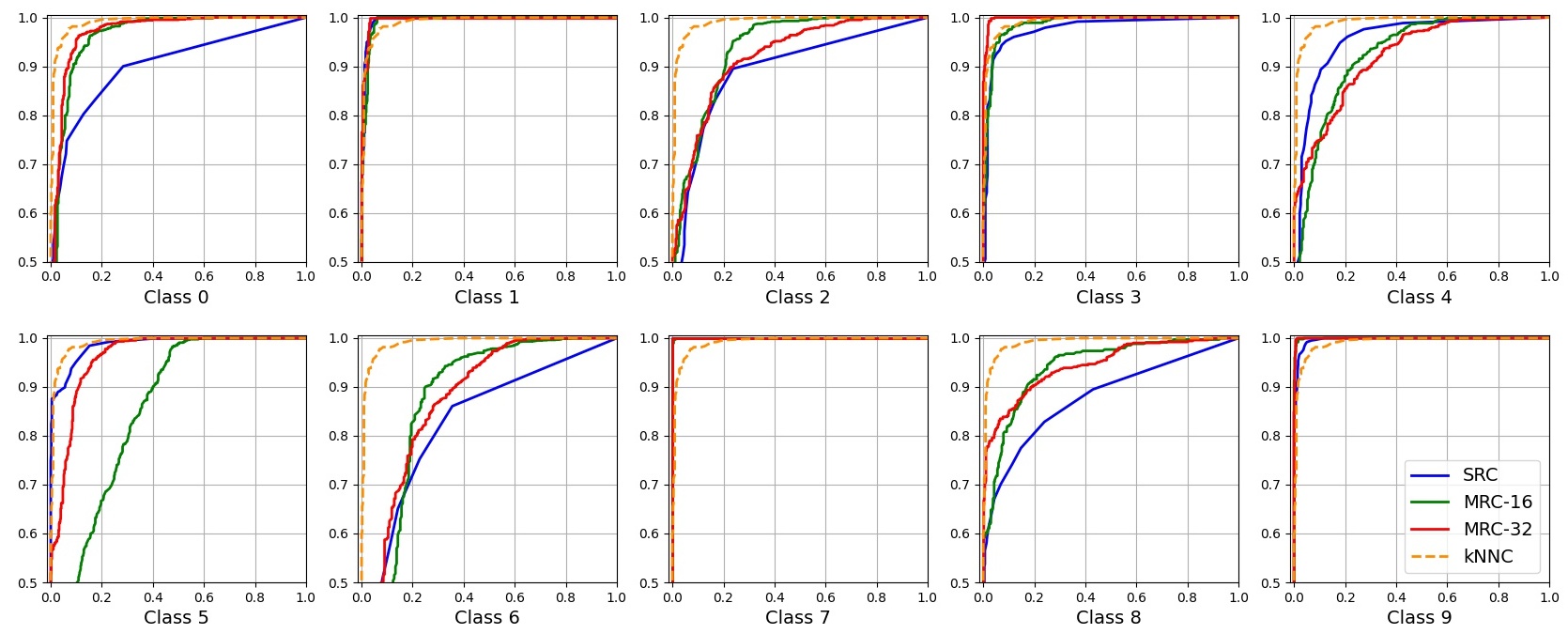}}
    \hfill
\caption{\small{ROC curves for each class of the MNIST dataset. They represent the TPR and FPR on the x-axis and y-axis, respectively. In each chart, the optimal threshold identifies the left-up corner that minimizes both FP and FN errors.}}
\end{figure*}

The calibration of thresholds has been performed using Receiver Operating Characteristic (ROC) analysis to compute 
the values $\tau_i$, for each class with index $i=1, \ldots, m$, that represent the best balance between 
minimizing the FPR and maximizing the TPR.
In this regard, portions of the Trusted Test Set and the Adversarial Set introduced in Section 7.2 of the paper have been used to test safe and unsafe inputs, 
respectively, during the calibration. In particular, 4500 samples from the Trusted Trusted Set have been used and the Adversarial Set has been built by generating unsafe inputs with all the methods reported in in the main manuscript (see Table 2 of the paper for the number of samples used for calibration). 
Figure \ref{fig:roc_mnist} reports the ROC curves (TPR and FPR on the x-axis and y-axis, respectively) obtained during the calibration 
for the MNIST dataset using the four CAMs.
The optimal balance between the TPR and the FPR is achieved at the top-left corner of the plots.
The same kind of results is reported in Figure~\ref{fig:roc_fmnist} for the F-MNIST dataset.

\subsection{Comparison with previous work}
Vision Guard \cite{vision_guard} and FeatureSqueezing \cite{feature_squeezing} are two runtime methods for detecting AEs. The fundamental assumption of these methods is that AEs are fragile inputs, i.e., they lose the adversarial strength if subject to pre-processing with image transformations. Thus, their key idea consists in processing one or more times a given input using transformations (e.g., jpeg-compression, pixel quantization, and smoothness filter), provide the corresponding images to the DNN of interest, and then compute the divergence among all obtained softmax outputs. This is accomplished by a certain formula: \textit{Kullback–Leibler divergence} for Vision-Guard and \textit{Norm-1} for FeatureSqueezing. The scalar value computed from such a formula represents how much the probability scores produced by the transformed versions disagree with the one obtained by the given input.
Whenever the value is larger than a certain threshold ($\tau_{VG}$ and $\tau_{FS}$ for VG e FS, respectively) the output is deemed an AEs. 

The following algorithms formalize Vision-Guard and Feature-Squeezing\footnote{In the original paper, the authors present multiple transformations for crafting transformed inputs. To provide a fair comparison, we applied the same transformations used by the authors on the tests performed with MNIST.}.

\begin{algorithm}[htb!]
 \SetAlgoLined
 \textbf{Input} Input image $x$\\
 \textbf{Output} \text{safe} or \text{unsafe} \\
 \algrule 
 $x_{jpg}$ = \text{JPEG\_compression}($x$) \\
 $out, ~out_{jpg} = f(x), ~f(x_{jpg})$ \\
 $J = min(D_{KL}(out, out_{jpg}), D_{KL}(out_{jpg}, out))$ \\
 \If{$J >= \tau_{VG}$}{
    \Return \text{unsafe}
 }
 \Return \text{safe}
\caption{Vision-Guard.}
\label{alg:vision-guard}
\end{algorithm}

\begin{algorithm}[htb!]
 \SetAlgoLined
 \textbf{Input} Input image $x$\\
 \textbf{Output} \text{safe} or \text{unsafe} \\
 \algrule 
 $x_{p}$ = \text{pixel\_quantization}($x$) \\
 $x_{s}$ = \text{smoothness\_filter}($x$) \\
 $out, ~ out_{s}, ~ out_{p} = f(x), ~f(x_{s}), ~f(x_{p})$ \\
 $J_p = |out - out_{p}|_1$ \\
 $J_s = |out - out_{s}|_1$ \\
 $J = max(J_p, J_s)$ \\
 \If{$J >= \tau_{FS}$}{
    \Return \text{unsafe}
 }
 \Return \text{safe}
\caption{FeatureSqueezing.}
\label{alg:features_squeezing}
\end{algorithm}

\textbf{Implementation details.} Both the methods require specific thresholds ($\tau_{FS}$ and $\tau_{VG})$ to determine if a given input is safe or not. Those thresholds need to be computed with a calibration based on a ROC analysis (in a similar manner as the one described above). 
To provide a fair comparison we used the same calibration set used to calibrate our methods).

\subsection{Relation between AEs and coverage metrics}

\begin{figure*}
\centering
  \subfloat[\label{fig_epsilon_cov}]{%
       \includegraphics[width=0.48\linewidth]{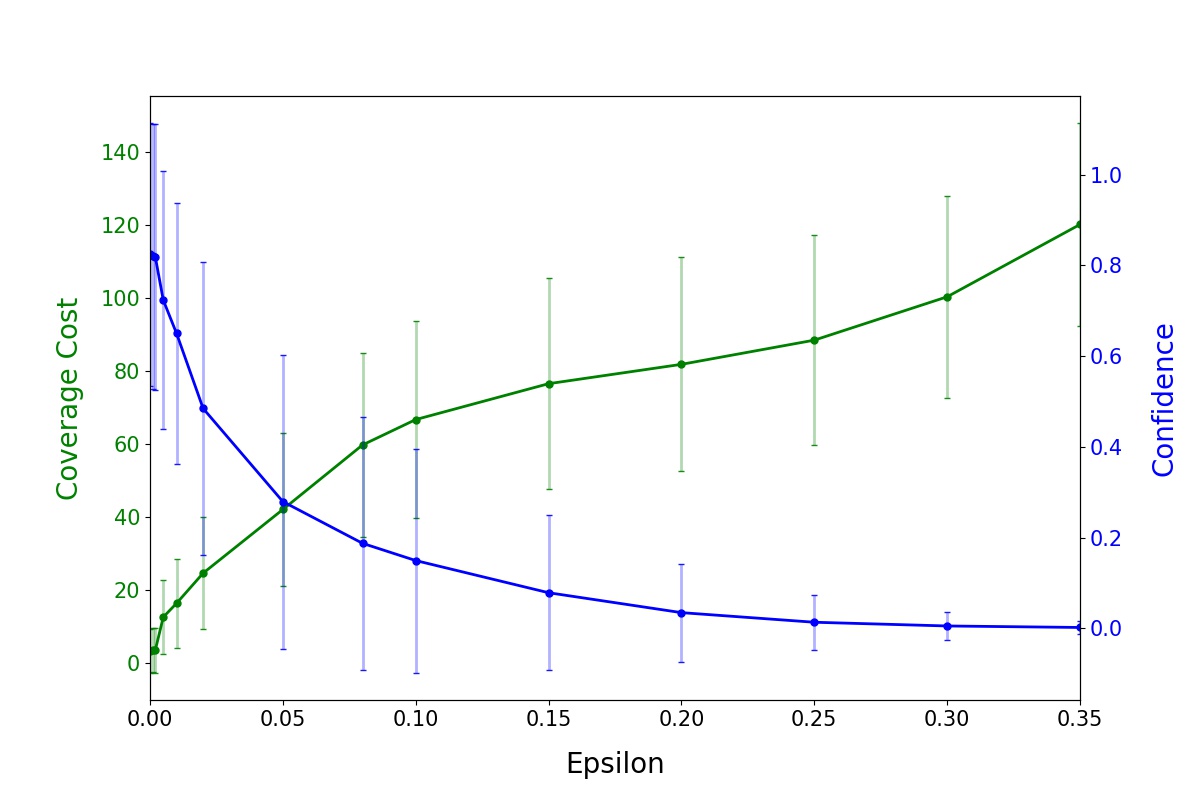}}
    \hfill
  \subfloat[\label{fig:num_aes_cov}]{%
        \includegraphics[width=0.43\linewidth]{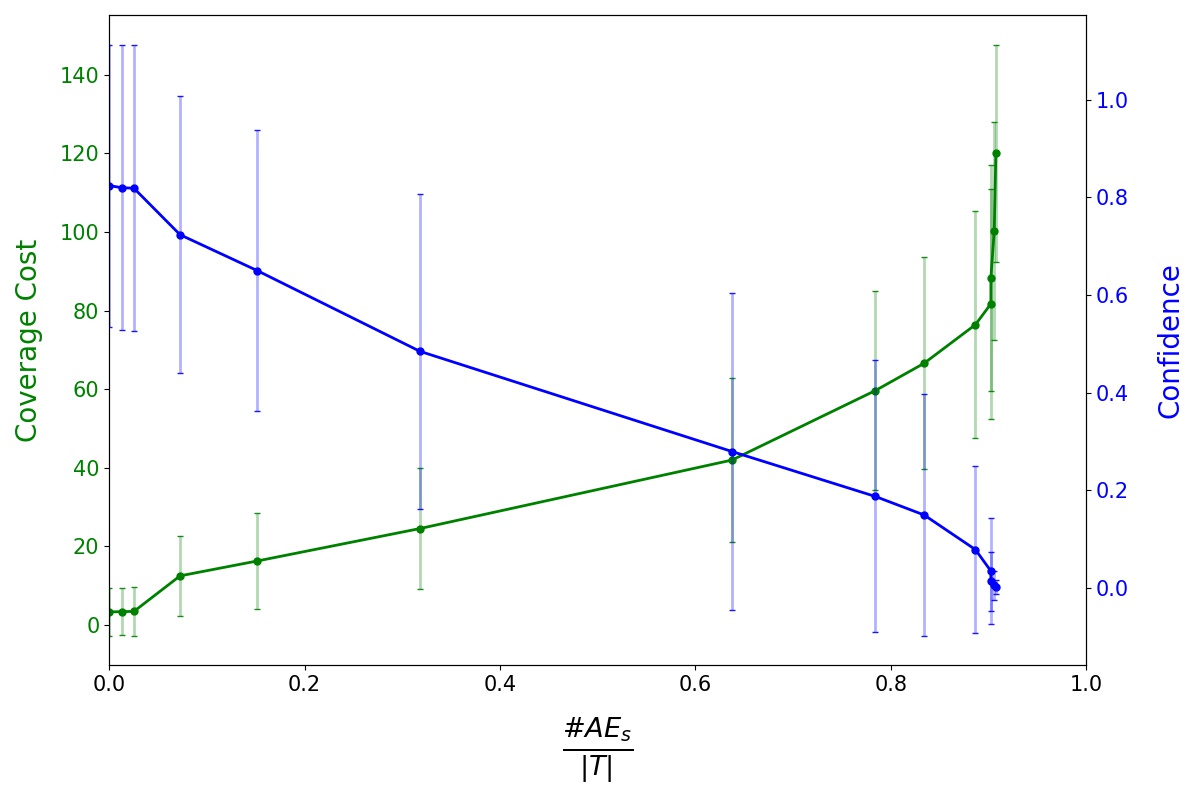}}
     \hfill
\caption{\small{Analysis of the Coverage cost and confidence score as a function of the $\epsilon$ parameter in the FGSM attack on F-MNIST (a) and on the corresponding number of AEs in the perturbed test set $\mathcal{T}$ (b).}}
\end{figure*}

To show the evident correlation between the adversarial perturbation and the proposed CAMs, we performed additional tests. The following experiments study the relationship between the magnitude of an applied perturbation (using the FGSM varying the $\epsilon$ on the original FMNIST test set) with the achieved \textit{mean confidence score} $\bar{c}$ and the corresponding \textit{mean coverage cost} $ \bar{\eta}$, which are both formally defined as follows:

\begin{equation}
\bar{c} = \frac{1}{|\mathcal{T}|} \sum_{x_i \in \mathcal{T}} c_i, \quad \quad
\bar{\eta} = \frac{1}{|\mathcal{T}|} \sum_{x_i \in \mathcal{T}} \eta_i,
\end{equation}
\noindent where $\mathcal{T}$ is the test set attacked with FGSM, while $c_i$ and $\eta_i$ are the confidence score and coverage cost corresponding to each attacked inputs $x_i$ in $\mathcal{T}$.

We evaluated the previous metrics among several attacked version of $T$, using FGSM with $\epsilon \in \{ 0.0, 0.01, 0.02, 0.05, 0.08, 0.1, 0.15, 0.2, 0.25, 0.3, 0.35 \}$. The results in Figure \ref{fig_epsilon_cov} show that the more the perturbation increases, the more the mean coverage cost $\bar{\eta}$ increases as well, hence decreasing the mean confidence accordingly. 

Figure \ref{fig:num_aes_cov} provides another important result. It plots the mean coverage cost as a function of the ratio of adversarial examples in the dataset, which is perturbed by varying $\epsilon$. Clearly, the larger the applied epsilon, the higher the number of generated adversarial examples in the attacked dataset.
These results show that both the number of adversarial examples and the mean coverage cost increase accordingly, meaning that the more unsafe a dataset is the higher its corresponding mean coverage cost is, and hence the lower the mean confidence score.

\end{document}